\documentclass{article}

\usepackage{microtype}
\usepackage{graphicx}
\usepackage{subcaption}
\usepackage{booktabs} %
\usepackage{multirow}
\usepackage{colortbl}
\usepackage[dvipsnames]{xcolor}

\usepackage{float}
\usepackage{placeins}
\usepackage{CJKutf8}
\usepackage{inconsolata}

\usepackage{stmaryrd}

\usepackage{url}

\usepackage{breakurl}
\usepackage[breaklinks]{hyperref}

\usepackage[accepted]{icml2025}

\usepackage{amsmath}
\usepackage{amssymb}
\usepackage{mathtools}
\usepackage{amsthm}
\usepackage{bbm}
\usepackage{makecell}
\usepackage{longtable}
\usepackage[capitalize,noabbrev]{cleveref}

\usepackage{pifont} %

\usepackage{xcolor}
\usepackage{color}
\usepackage{tcolorbox}
\tcbuselibrary{listings,breakable}

\usepackage{titlesec}
\titlespacing*{\paragraph}{\parindent}{0.25ex}{1ex}

\theoremstyle{plain}

\theoremstyle{definition}

\theoremstyle{remark}

\usepackage[textsize=tiny]{todonotes}

\definecolor{x11purple}{RGB}{160, 32, 240}
\newcommand{\tohigh}{\tau}

\newcommand{\highlevel}{\mathcal{H}}
\newcommand{\lowlevel}{\mathcal{M}}

\newcommand{\verified}{\mathbb{V}}

\newcommand{\var}{\mathcal{V}}
\newcommand{\inputvar}{\mathcal{X}}
\newcommand{\outputvar}{\mathcal{Y}}
\newcommand{\getval}{\texttt{GetVal}}
\newcommand{\onehot}{\mathbbm{1}}

\icmltitlerunning{Internal Causal Mechanisms Robustly Predict Language Model Out-of-Distribution Behaviors}

\begin{document}

\twocolumn[
\icmltitle{Internal Causal Mechanisms Robustly Predict \\ Language Model Out-of-Distribution Behaviors}

\icmlsetsymbol{equal}{*}

\begin{icmlauthorlist}
\icmlauthor{Jing Huang}{equal,stanford}
\icmlauthor{Junyi Tao}{equal,stanford}  
\icmlauthor{Thomas Icard}{stanford} 
\icmlauthor{Diyi Yang}{stanford}    
\icmlauthor{Christopher Potts}{stanford}
\end{icmlauthorlist}
\icmlaffiliation{stanford}{Stanford University}
\icmlcorrespondingauthor{Jing Huang}{hij@stanford.edu}

\vskip 0.3in
]

\printAffiliationsAndNotice{\icmlEqualContribution} %

\begin{abstract}
Interpretability research now offers a variety of techniques for identifying
abstract internal mechanisms in neural networks.
Can such techniques be used to predict how models will behave on out-of-distribution examples? In this work, we provide a positive answer to this question.
Through a diverse set of language modeling tasks---including symbol manipulation, knowledge retrieval, and instruction following---we show that the most robust features for correctness prediction are those that play a distinctive causal role in the model's behavior. Specifically, we propose two methods that leverage causal mechanisms to predict the correctness of model outputs:
\emph{counterfactual simulation} (checking whether key causal variables are realized) and \emph{value probing} (using the values of those variables to make predictions).
Both achieve high AUC-ROC in distribution and outperform methods that rely on causal-agnostic features in out-of-distribution settings, where predicting model behaviors is more crucial. Our work thus highlights a novel and significant application for internal causal analysis of language models.

\end{abstract}

\section{Introduction}

A core task in interpretability research is to identify internal mechanisms that mediate a model's input--output behavior. The motivations for this focus are several. One is simply that compressed, algorithmic descriptions of complex models can help confer a sense of understanding insofar as the algorithms themselves are relatively simple and intuitive. Another, related motivation stems from an intuition that the very presence of abstract algorithmic mechanisms is deeply related to a model's ability to \textit{generalize} beyond the specific domains in which it has been trained. Indeed, it is commonly assumed that a model will succeed at a generalization task if and only if it has induced a mechanism that implements a ``correct" algorithm for that task (provided one exists).

There are two directions to this hypothesized correspondence. In one direction, when a model is successful, this warrants a search for the internal mechanisms that explain success. That is, we search for an appropriate \textit{abstraction} of the model to elucidate how the model was able to generalize~\cite{geiger2024causalabstraction}. 
This is the focus of interpretability work that aims to understand the internal causal mechanisms that mediate LLM behaviors~\cite{vig2020gender,geiger2021causal, finlayson-etal-2021-causal,olsson2022context,wang2023ioi, meng2023massediting,wu2023interpretabilityatscale}. 

\begin{figure}[t]
    \vspace{0.5ex}
    \centering
   \resizebox{0.90\columnwidth}{!}{
    \includegraphics{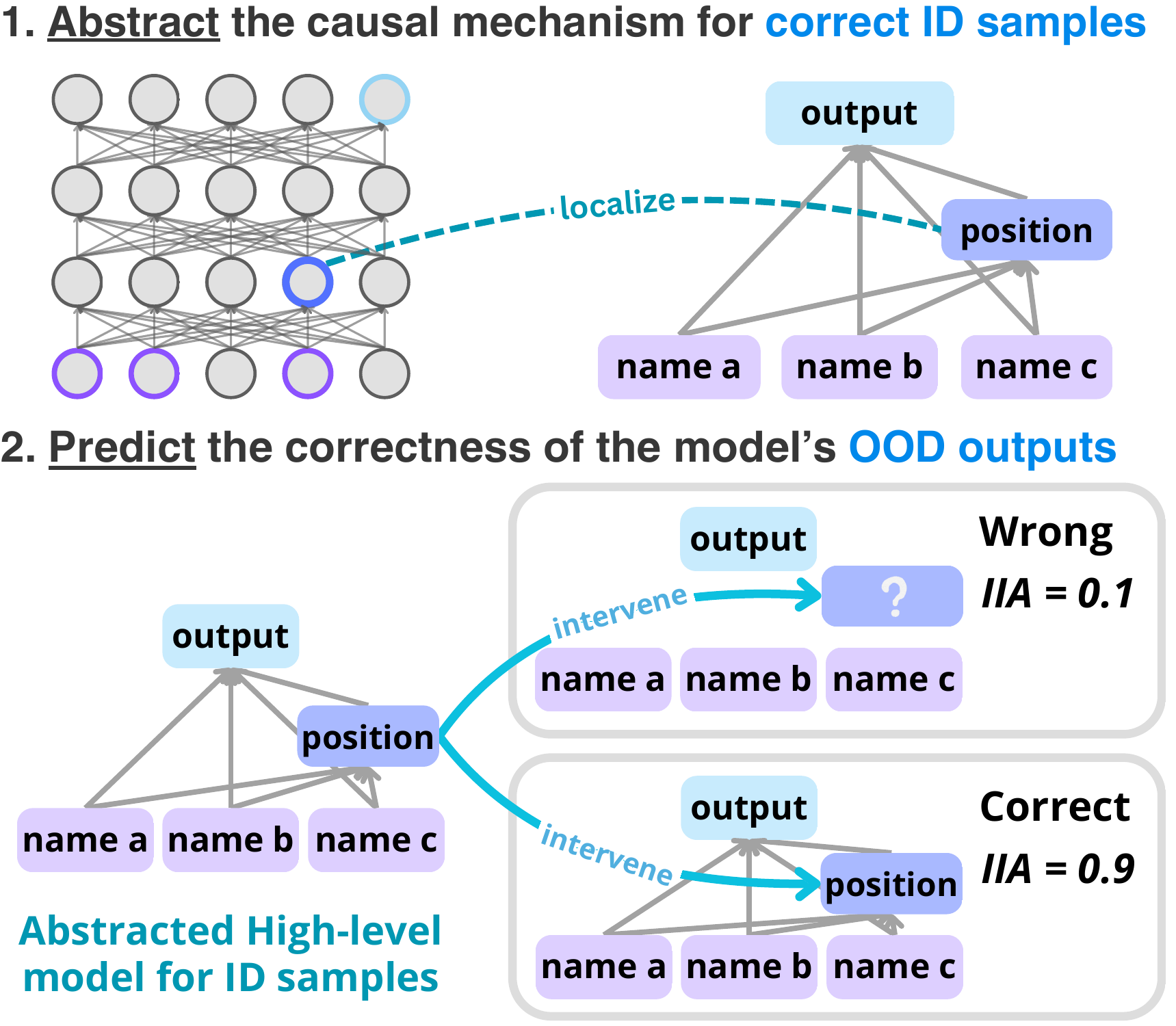}
    }
    \vspace{-1ex}
    \caption{We propose a two-stage framework to predict correctness of model outputs under distribution shifts, illustrated using the counterfactual simulation method on the Indirect Object Identification task~\cite{wang2023ioi}. \textbf{Stage 1 (Abstract)}: Identify the abstract causal mechanisms the model uses to solve the task correctly on in-distribution data. \textbf{Stage 2 (Predict)}: Predict the model's output correctness on out-of-distribution inputs by checking whether it implements the same mechanisms. \textcolor{RoyalPurple}{\textbf{Dark blue}} indicates the key causal variables used for correctness prediction.
    }
    \label{fig:overview}
    \vspace{-2ex}
\end{figure}

It would be desirable if we were able to move in the other direction as well, from identifying an abstract mechanism to better predicting model generalization behaviors.
We refer to this direction as \textit{prediction}.
In this work, we explore a specific instance of the prediction direction---predicting the correctness of LLM behaviors under distribution shifts.

As language models are increasingly deployed in high-stakes, open-ended scenarios, it is critical that we have a handle on how to prevent incorrect or unsafe outputs across a wide range of unanticipated user inputs. However, behavioral testing usually cannot be comprehensive given the compositionally large input space of a natural language task and the potential for distribution shifts across deployed environments. 
The traditional approach is to estimate the correctness of model predictions using confidence scores~\cite{hendrycks2017a,lakshminarayanan2017ensembles,guo2017calibration}. 
A more recent line of work uses internal representations of the model (usually supervised probes or use other statistical measures of model activations)
to predict the model's correctness on factual retrieval tasks~\cite{azaria-mitchell-2023-internal,zou2023representationengineering,orgad2024llmsknowshowintrinsic,marks2024the,gottesman-geva-2024-estimating,ferrando2024knowledgeawareness,ji-etal-2024-llm,yin2024truthfulness,chuang-etal-2024-lookback}.

With the present paper, we seek to show that causal interpretability methods can play a critical role in correctness prediction. Our central question is the following: if we know what internal features causally contribute to the model's predictions in the in-distribution setting, can we use them as robust predictors of the model's out-of-distribution behaviors?
To answer this question, we propose two methods that leverage previously identified causal mechanisms to predict model generalization behavior (as illustrated in Figure~\ref{fig:overview}). 
The first method is \textit{counterfactual simulation}, which predicts whether the model output is correct by checking whether it computes the key causal variables needed for solving the task.
The second method is \textit{value probing}, which predicts correctness based on the decision boundaries of different values taken by a causal variable.

We compare our proposed methods against existing methods across a diverse set of five tasks under in-distribution (ID) and multiple out-of-distribution (OOD) settings. 
The baseline methods include traditional confidence score-based methods and causal-agnostic probing methods using internal activations.  
The evaluated tasks range from highly curated symbolic tasks to more open-ended tasks where only partial mechanisms have been identified, including Indirect Object Identification \cite{wang2023ioi}, PriceTag \cite{wu2023interpretabilityatscale}, RAVEL \cite{huang-etal-2024-ravel}, MMLU \cite{hendrycks2021measuringmassivemultitasklanguage}, and  UnlearnHP \cite{thaker2024guardrailbaselinesunlearningllms}. 
Our results show that methods leveraging task-relevant causal features achieve high AUC-ROC scores under ID settings---comparable to or slightly better than the confidence score baseline. Crucially, these causal feature-based methods significantly outperform existing baselines in OOD settings. Our strongest method, counterfactual simulation, improves average AUC-ROC by 13.84\% over prior baselines.

Overall, our results show that internal causal features are more robust predictors of correctness than non-causal ones, especially under distribution shifts. More broadly, our findings highlight a novel and important application of causal interpretability analysis, suggesting a path from understanding model internals to improving model safety and reliability.

\section{Background}

\subsection{Finding Abstract Causal Mechanisms in LLMs}
\label{sec:background_localization}

A trained neural network can be construed as a causal model, with variables being the neurons in the network and the functional mechanisms being given by weight matrix multiplications across the layers that comprise the neural network. Causal interpretability research asks the question: Is there a \textit{more abstract} causal model that adequately captures the structure of the neural network when it comes to a particular task that the network successfully performs ~\cite{geiger2021causal,geiger2024causalabstraction}? This line of work is built on \textit{causal abstraction}, which is concerned with the general question of when one causal model can be said to \textit{abstract} another~\cite{RWB+17}. In the interpretability setting, the least abstract causal model is just the neural network itself, and the more abstract causal model is referred to as \textit{high-level} model.
Below we review the workflow of finding such abstractions using the language of causal abstraction.

\paragraph{Proposing high-level causal model} High-level causal models should capture the shared structure among examples of the task, such as the underlying algorithmic description of the task solution. Such models are typically proposed by researchers based on prior knowledge of the task. Some recent work attempts to reduce reliance on human priors by automating this process~\cite{hernandez2022natural,bills2023language,sun2025hyperdas}. In this work, we adopt high-level models that have been proposed and verified in the literature.

\paragraph{Localizing high-level variables in neural networks} Given a high-level model, we search for its implementation in the neural network. Existing methods  localize each high-level variable to the model representation space through a bijective function~\cite{huang-etal-2024-ravel,geiger2024causalabstraction}.  

We focus on one such method, Distributed Alignment Search (DAS; \citealt{geiger2024das}), which achieves the highest accuracy on localization benchmarks~\cite{huang-etal-2024-ravel,arora-etal-2024-causalgym,mueller2025mib}. DAS uses \textit{distributed interchange interventions} to identify an orthonormal basis $Q$ as the bijective function. Let $\texttt{GetVal}(\lowlevel(x), R)$ be the value of some model representation $R$ when the model $\lowlevel$ processes an input $x$, and let $\lowlevel_{R\leftarrow r}(x)$ be the output of $\lowlevel$ when it processes $x$, but with $R$ set to value $r$. Given a pair of inputs $x_{\textit{base,i}}$ and $x_{\textit{src,i}}$, we perform an intervention on the subspace that localizes the variable:
\begin{align}
    r_{\textit{base,i}} &= \texttt{GetVal}(\lowlevel(x_{\textit{base,i}}), R) \nonumber \\
    r_{\textit{src,i}} &= \texttt{GetVal}(\lowlevel(x_{\textit{src,i}}), R) \nonumber \\
    r_{\textit{inv,i}} &= (I - Q^\top Q)\, r_{\textit{base,i}} + Q^\top Q \, r_{\textit{src,i}} \label{eq:subspace_localization}\tag{II}
\end{align}
To find the optimal $Q$, DAS minimizes loss $\ell_{Q}$ over $N$ counterfactual pairs $D_{\textit{cf}}=\{(x_{\textit{base,i}}, x_{\textit{src,i}}, y_{cf,i})\}_{i=1}^N$, i.e., a pair of randomly sampled inputs and a counterfactual label determined by the high-level model, as shown in Eq~\ref{eq:subspace_localization}.
\begin{align}
    \textbf{Distri}&\textbf{buted Alignment Search} \nonumber \\
    \ell_{Q} &= \mathbb{E}_{1\leq i\leq N}\big[-y_{\textit{cf,i}} \cdot \log y_{\textit{inv,i}}\big] \nonumber \\
    \text{where} \quad y_{\textit{inv,i}} &= \lowlevel_{R \leftarrow r_{\textit{inv,i}}}(x_{\textit{base,i}}) 
 \label{eq:das}\tag{DAS}
\end{align}
We define the metric \textit{interchange intervention accuracy} (IIA) to measure the extent to which an implementation exists in $\lowlevel$:
\begin{equation}
    \texttt{IIA}(D_{\textit{cf}}) = \mathop{\mathbb{E}}_{1\leq i \leq N}\big[\tau(y_{cf,i})=\tau(y_{inv,i})\big]
    \label{eq:iia}\tag{IIA}
\end{equation}
where $\tau$ maps the low-level representation to a high-level value, e.g., decoding logits into a token.

\subsection{Correctness Estimation}
Correctness estimation is a fundamental problem in machine learning evaluation and reliability assessment. 
It is the problem of predicting whether a model output matches the ground truth. For a classifier, a natural choice is to use  probabilistic predictions, i.e., the confidence scores~\cite{hendrycks2017a,lakshminarayanan2017ensembles}. However, LMs' outputs are structured predictions which add complexities to calibration~\cite{kuleshov2015calibrated}, and deep neural models are only well-calibrated to in-domain inputs after temperature scaling~\cite{guo2017calibration,desai-durrett-2020-calibration,kadavath2022languagemodelsmostlyknow,openai2024gpt4technicalreport,wang-etal-2024-answer-c}. Alternatives have been proposed, such as using verbalized confidence scores for instruction-tuned models~\cite{kadavath2022languagemodelsmostlyknow,tian-etal-2023-just}. Most related to ours is the line of work using internal representations to estimate the knowledge stored in LMs~\cite{gottesman-geva-2024-estimating,marks2024the,ferrando2024knowledgeawareness}, assess the risk of hallucination~\cite{azaria-mitchell-2023-internal,zou2023representationengineering,du2024haloscope,chuang-etal-2024-lookback,ji-etal-2024-llm,yin2024truthfulness,orgad2024llmsknowshowintrinsic}, and predict jailbreaking behaviors~\cite{arditi2024refusal}. However, identifying robust features from internal representations that generalize across tasks and datasets remains a challenge~\cite{orgad2024llmsknowshowintrinsic}. This is the problem that we focus on in this work.

\section{Problem Setting}

\label{sec:correctness_pred_task}

We are interested in predicting whether a language model $\lowlevel$ produces correct answers on a task under distribution shifts. We formalize this as a correctness prediction problem.

\paragraph{Task} We first define our task as a function that takes as the input a sequence of variables $\inputvar_1, \dots, \inputvar_n$ and returns an output \(\outputvar\). %
For symbolic tasks, the function is the symbolic algorithm; for knowledge retrieval tasks, the function is simply a look-up table, mapping each input to an output. 
Next, we define a prompt template \( T \) and use it to generate all valid prompts by assigning values to the input variables. We focus on tasks whose correct outputs are completely determined by processing the input variables using the defined task function. Everything else in the template, besides the input variables, serves merely as background conditions.

\begin{figure}[t]
    \centering
   \resizebox{1\columnwidth}{!}{
    \includegraphics{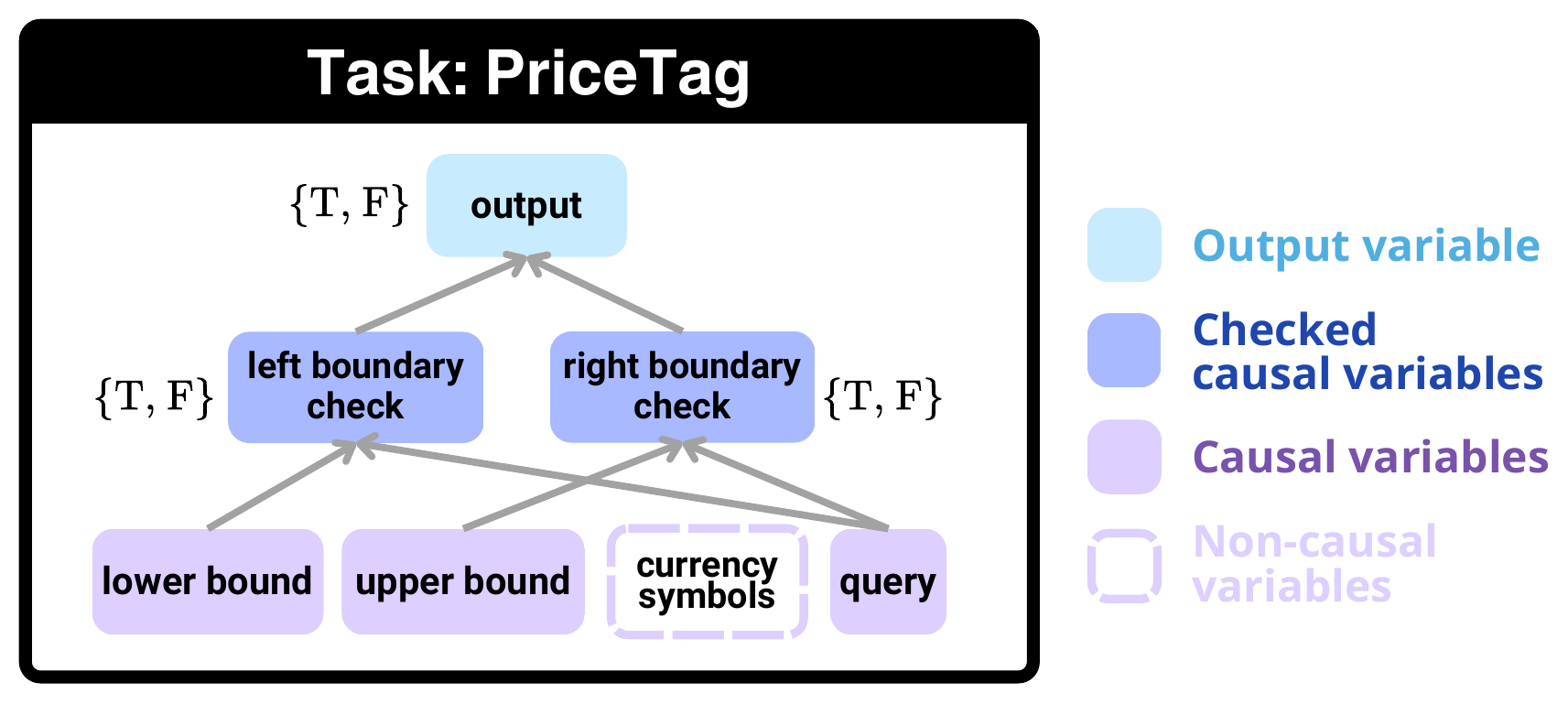}
    }
    \vspace{-5ex}
    \caption{\textbf{The identified high-level model for the PriceTag task.} See high-level models for all tasks in Figure  \ref{fig:highlevel_models}.}
    \label{fig:pricetag_highlevel_model}
    \vspace{-1ex}
\end{figure}

Take the PriceTag task as an example (illustrated in Figure~\ref{fig:pricetag_highlevel_model}). 
 The input variables are $\inputvar_\textit{lb}$, $\inputvar_{ub}$, and $\inputvar_{q}$, representing the lower bound, the upper bound, and the query. The task function is given by $f(\inputvar_{\textit{ub}}, \inputvar_{\textit{lb}}, \inputvar_{\textit{q}}) = \llbracket \inputvar_{\textit{ub}} > \inputvar_{\textit{q}} \rrbracket \land \llbracket \inputvar_{\textit{lb}} < \inputvar_{\textit{q}} \rrbracket$.
We use the template $T = $ ``Does the following item cost between \texttt{\$\{lower\_bound\}} and \texttt{\$\{upper\_bound\}}? Item: \texttt{\$\{query\}}." to generate all prompts, where each input variable is assigned a number sampled from [0, 20].

\paragraph{ID vs.~OOD setting} For each task, we construct one in-distribution and multiple OOD settings. For the in-distribution setting, we fill the input variables and the template with a set of values frequent in pre-training corpora. For the OOD settings, we either introduce perturbations to background conditions irrelevant to solving the task (e.g., replacing the dollar sign with a lira sign in PriceTag) or change the values of input variables (e.g., changing the language of names in IOI). These changes empirically lead to a decline in task accuracy compared with the in-distribution setting.
The assumption is that if the task function remains the same, a model employing a systematic solution should continue to make correct predictions despite irrelevant changes. 

\paragraph{Problem formulation}
Given a task input $x_i$, we compare the model's prediction \(\lowlevel(x_i)\) to the ground-truth label, which yields a correctness label \(l_i \in \{0,1\}\), where \(1\) indicates correct. The resulting dataset \(D  = \{(x_i, \lowlevel(x_i), l_i)\}_{i=1}^N\) consists of $N$ tuples of inputs, model predictions, and correctness labels.
As most prediction methods are supervised, we additionally assume access to a labeled subset of $M$ ID examples which we call the \textit{verified set}: $\verified = \{(x_i, \lowlevel(x_i), l_i)\}_{i=1}^M$. 
At test time, we evaluate our methods on both in-distribution and OOD settings. To do this, we sample unlabeled test inputs \(x_{\textit{test}}\) from the ID dataset (excluding the verified set) and the OOD datasets. All samples are drawn independently and identically.
The correctness prediction problem is thus formulated as a binary classification task predicting \(P(l = 1 \mid \lowlevel, x_{\textit{test}})\).

\section{Methods}

In this section, we first review existing methods and then introduce two new methods that leverage internal causal mechanisms to predict correctness. We categorize these methods into three families based on the types of features they use: (1) output probabilities, (2) \textit{internal causal-agnostic features} that are not informed by causal understandings of the model's internal mechanisms, and (3) \textit{internal causal features} that correspond to identified abstract causal variables.

\subsection{Output Probabilities}
\label{sec:output_probabilities}
The first family of methods uses output probabilities as features for predicting output correctness, essentially treating them as confidence scores. The predictive power of confidence scores is grounded in the modeling objective---in a perfectly calibrated model, the predicted probability is exactly equal to the expected accuracy. Since LMs are not always perfectly calibrated, we additionally apply temperature scaling to confidence scores for a fairer evaluation of their predictiveness~\cite{openai2024gpt4technicalreport,kadavath2022languagemodelsmostlyknow}.

Existing work typically aggregates token-level confidence scores to predict the correctness of an output sequence. 
Given a test input \(x_{\textit{test}}\), let \(t_1,\dots, t_n\) be the sequence of output tokens from \(\lowlevel\). Denote the confidence score \(c_i\) as the probability of token \(t_i\) at decoding step \(i\), computed as logits divided by the temperature parameter \(T\).
The correctness \(f(\lowlevel, x_{\textit{test}})\) is approximated as the log-likelihood of a subset of output tokens $S$, normalized by its size.
\begin{align}
    \textbf{Confidence }&\textbf{Scores}
    \nonumber\\
    f(\lowlevel, x_{\textit{test}}) = \frac{1}{|S|}\log\Bigg(\prod_{c_i\in S} &c_i\Bigg) = \frac{1}{|S|}\sum_{c_i\in S} \log\left(c_i\right)
    \label{eq:conf_method}
\end{align}
There are two common variations that differ by how they select the set of tokens in $S$. The simplest one,  \textbf{First n Tokens}, uses the first $n$ decoded tokens, where $n$ is a hyperparameter. An alternative is to select tokens that represent the actual answer, which we refer to as \textbf{Answer Tokens}.

Since AUC-ROC evaluation is ranking-based, we can directly use the log-likelihood without further normalization.

\subsection{Internal Causal-Agnostic Features}
\label{sec:causal_agnostic}
Rather than directly using output probabilities, the second family of methods extracts internal features as signals of output correctness. A typical approach is to train linear probes on extracted features using correctness labels as supervision. The choice of \textit{which} internal representations to use usually relies on heuristics---since it is not informed by a causal understanding of how the model solves the task, we refer to these extracted features as \textit{causal-agnostic}. Commonly used heuristics include location-based selection, such as selecting features of the last prompt token and the last model layer~\cite{zou2023representationengineering,yin2024truthfulness}. In principle, probing methods can apply to internal features at any location. However, without an understanding of the features’ functional roles, these methods cannot distinguish causal features from others, nor can they systematically identify features robust under distribution shifts. Instead, they rely solely on observed probing accuracy on the available data.

Formally, given an internal feature $x_{L}$ at location $L$, a correctness label $l\in[0, 1]$, and a probe $\phi=\sigma(Wx_{L} + b)$, where $\sigma$ is the sigmoid function, the training objective is:
\begin{equation}
\begin{aligned}
    \ell_{W,b} = \mathbb{E}\big[-l\cdot\log(\phi(x_{L}))]
    \label{eq:correctness_probe_train}  
\end{aligned}
\end{equation}
At inference time, correctness is predicted as follows:
\begin{align}
    \textbf{Correctness Probing }&\textbf{Using Features $x_L$} \nonumber \\
     f(\lowlevel, x_{\textit{test}}) &= \phi(x_{L})
    \label{eq:probex_method}
\end{align}

\subsection{Internal Causal Features}
\label{sec:causal}
We propose a third family of methods that leverage causal features for correctness prediction.
In contrast to prior methods, we look specifically for internal \textit{causal} features that mediate the model’s input--output behavior relevant to the task, guided by previously discovered high-level causal mechanisms. 
We introduce two methods: (1) counterfactual simulation, which identifies causal features robust to counterfactual permutations of inputs; and (2) value probing, which measures the distance between model predictions and learned decision boundaries associated with possible values of a causal variable.

\paragraph{Counterfactual simulation}

Our first method is built on the assumption that if a model can consistently make the correct predictions, the model must implement a systematic solution. Suppose we have identified such a solution (represented by an abstract high-level causal model $\highlevel$)  for all seen examples that are predicted correctly. We then expect the model to continue using the same solution to correctly predict unseen examples of the same task. If, however, the model no longer implements this solution, its prediction is unlikely to have arisen as part of the systematic solution that it found for other inputs. With this in mind, we want to check %
whether the model successfully computes the key causal variables specified in $\highlevel$.

Given a task and a model $\lowlevel$ of interest, counterfactual simulation takes as input the identified causal mechanisms (as discussed in Section~\ref{sec:background_localization}), comprising a high-level causal model $\highlevel$ and its corresponding neural representations.

We illustrate using a simple-yet-common chain of three variables as a {running example}: $\highlevel: \inputvar\rightarrow\var\rightarrow\outputvar$, with localized neural representations $X$, $V$, and $Y$ (e.g., $V$ could be a residual subspace at layer 3's last token).
We predict output correctness by estimating $P(Y|V)$.
Ideally, if the model implements the correct algorithm, the output $Y$ depends more generally only on its immediate causal parents in the identified high-level causal model, which is in this case: $P(Y=y|X=x)=P(Y=y|V=\getval(\lowlevel(x), V))$. 
Yet in practice, the output $Y$ is also affected by other aspects of the input $X$ that do not feed into $V$ (and hence live outside of~$\highlevel$) to various extents. We term these task-irrelevant aspects \textit{background variables} and denote them as $B=V^{\perp}$, the orthogonal complement of the identified representation of intermediate causal variables. 

The distribution $P(Y|V)$ can be directly estimated using outputs from localization algorithms like DAS. In fact, the output $y_{\textit{inv,i}}$ in Eq.~\ref{eq:das} can be interpreted as estimating $P(Y|V, B)$: Given a base input $x_{\textit{base}}$ in the verified set $\mathbb{V}$, a test input $x_{\textit{src}}$, and a counterfactual label $y_{\textit{cf}}=\onehot(\tohigh(\lowlevel(x_{\textit{src}})))$, where $\tohigh$ decodes the output distribution into a token and $\onehot$ maps the output token to a one-hot encoding, let  $v_\textit{src} =\getval(\lowlevel(x_\textit{src}), V)$, $b_\textit{base}=\getval(\lowlevel(x_\textit{base}), B)$, and $y_{\textit{inv}}=\lowlevel_{V\leftarrow v_\textit{src}}(x_\textit{base})$. The value of $y_{\textit{cf}}\cdot y_{\textit{inv}}$ then estimates $P(Y=y_{\textit{cf}}|V=v_{\textit{src}}, B=b_{\textit{base}})$. 
Since the values of $V$ and $B$ are drawn independently, we can estimate $P(Y|V)$ by marginalizing over the background variable $B$:
\begin{eqnarray}
P(Y|V) &=& \sum_{B} P(Y|V,B)P(B|V) \nonumber \\
&=& \mathbb{E}_{B}\big[P(Y|V,B)\big]  \label{eq:intervention_prob}
\end{eqnarray}
Eq. \ref{eq:intervention_prob} suggests that for a test input, we can approximate $P(Y|V)$ by sampling multiple values of $B$ from inputs in the verified set $\mathbb{V}$. This can also be interpreted as measuring the robustness of the causal relation between $\mathcal{X}$ and $\mathcal{Y}$ against perturbations in the background variables, with the expectation that more robust causal relations are more likely to produce correct predictions under distribution shifts.

There are several ways to measure the success of counterfactual simulation.  One is to use the probability of the first $n$ decoded tokens, which is simple and effective as an optimization objective for localization. Given $k$ samples from $\verified$ and an output of $n$ tokens, the correctness prediction is:
\begin{equation}
\begin{aligned}
    \textbf{Counterfactual Simulation (First n Tokens)} \\
     f(\lowlevel, x_{\textit{test}}) = \frac{1}{kn}\sum_{i=1}^{k}\sum_{t=1}^{n}-y_{\textit{cf},t}\log(y_{\textit{inv},t})
     \label{eq:intervention_method}
\end{aligned}
\end{equation}

A more accurate measure would be to directly estimate the probability of certain target concepts that indicate the correct answer (e.g., certain proper nouns), if the outputs are well-structured or we know how to parse them.  This means introducing a mapping $\tau_{\textit{out}}$ that maps the output sequence $y_{\textit{inv}}$ to some value of the corresponding high-level output variable $\tau_{\textit{out}}(y_{\textit{cf}})$:
\begin{equation}
\begin{aligned}
    \textbf{Counterfactual Simulation (Output Mapping)} \\
     f(\lowlevel, x_{\textit{test}}) = \frac{1}{k}\sum_{i=1}^{k}(\tau_{\textit{out}}(y_{\textit{cf}}) = \tau_{\textit{out}}(y_{\textit{inv}}))
     \label{eq:intervention_method_mapping}
\end{aligned}
\end{equation}
Note that defining this mapping typically demands prior knowledge of the task (e.g., which token indicates correctness) and output parsing rules (e.g., regex patterns), and thus may not be readily available for all tasks.

We generalize counterfactual simulation to tasks with more complex high-level models.  To do this, we verify whether the model correctly computes each causal variable individually, treating all other variables as background variables. Rather than directly using the test example's output as the counterfactual label $y_\textit{cf}$, we derive it by evaluating the high-level model under an intervention that sets $\var$ to its value in the test example. Take the IOI task as an example (the high-level model is shown in Figure~\ref{fig:overview}). Given a verified example with three input variables $\mathcal{X}_a=\bar{x}_a, \mathcal{X}_b=\bar{x}_b, \mathcal{X}_c=\bar{x}_c$, we check whether the model computes the position variable~$\mathcal{V}$ correctly for a test input. The correct counterfactual label is given by $y_\textit{cf}=\onehot(\bar{x}_a$ if $\bar{v}=\texttt{first}$ else $\bar{x}_b)$, where $\bar{v}\in\{\texttt{first}, \texttt{second}\}$ is inferred from either the source input or the test example output.

\paragraph{Value probing} Although counterfactual simulation does not require additional training, it still needs decoding model outputs and has a computation overhead from running $k$ forward passes, where $k$ is the number of samples. To avoid this, once we have localized the causal variable $\var$, we ask whether it is possible to project the decision boundaries of $P(Y|V)$ into the subspace that encodes $\var$ and compute the projected decision boundaries efficiently.

Since the representation of $\var$ encodes the minimal set of features necessary for counterfactual predictions, a natural strategy is to find a set of local decision boundaries separating distinct values of $\var$. For a language model $\lowlevel$, we assume the values that $\var$ can take are discrete. Let $\mathbb{D_{\var\mapsto\mathcal{Y}}}=\{\bar{v}_{1}, \dots, \bar{v}_{m}\}$ be the set of all values taken by $\var$ when the causal relation between $\var$ and $\mathcal{Y}$ holds. Let $\tau$ be a predictor that takes $\var$ as input and estimates the probability of $\mathcal{V} = \bar{v}_{i}$ on the verified set distribution. The predictor $\tau$ can be learned with a standard classification algorithm.

While $\tau$ could be any class of models, a particularly useful class is linear models. For such a linear model to exist, the representation of $\var$ must follow a linear structure: the representations of $\var$ lie in a linear subspace and distinct values of $\var$ are linearly separable. We parameterize the linear model $\tau$ by $W_{\tau}$ and learn it via standard classification:
\begin{equation}
\begin{aligned}
    \ell_{W_\tau} = \mathop{\mathbb{E}}_{x \in \mathbb{V}}\big[-\onehot(\bar{v})\cdot\log(\tau(x))]
    \label{eq:value_probe}  
\end{aligned}
\end{equation}
To predict output correctness, We use the probability of the most likely class of $\tau$:
\begin{equation}
\begin{aligned}
 \textbf{Value Probing Using Features $x$}\quad \\
 f(\lowlevel, x_{\textit{test}}) = \max_{1\leq i\leq m}\{\tau(x)_i\}
\end{aligned}
\end{equation}
The linear structure suggests two potential sources of low-confidence predictions: (1) when representations fall between classes, i.e., close to local decision boundaries; and (2) when representations fall outside the region enclosed by all classes, i.e., extrapolation errors. We empirically investigate whether these low-confidence predictions correspond to incorrect model predictions in Section~\ref{sec:exp}.

\label{sec:exp}

\section{Experiments}
\begin{table}[t!]
\setlength{\tabcolsep}{4pt}
    \setlength\extrarowheight{4pt}
    \centering
    \resizebox{1\columnwidth}{!}{
    \begin{tabular}{>{\raggedright\arraybackslash}p{2.2cm} ccccc}
        \toprule
        \footnotesize \textbf{Method} &
        \footnotesize \makecell{\textbf{Requires} \\ \textbf{Training}} &
        \footnotesize \makecell{\textbf{Requires} \\ \textbf{Wrong Samples}} &
        \footnotesize \makecell{\textbf{Requires} \\ 
        \textbf{Counterfactuals}} &
        \footnotesize \makecell{\textbf{Requires} \\ \textbf{Decoding}} &
        \footnotesize \makecell{\textbf{Inference} \\ \textbf{Cost}} \\
        \midrule
        Confidence Score & \ding{55} & \ding{55} & \ding{55} & \ding{51} & $\times N$ \\
        Correctness Probing & \ding{51} & \ding{51} & \ding{55} & \textit{Maybe} & $\times 0.5$--$N$ \\
        \addlinespace
        \makecell[l]{Counterfactual \\ Simulation} & \makecell{\textit{Localization}\\ \textit{Only}} & \ding{55} & \ding{51} & \ding{51} & $\times NK$ \\
        Value Probing & \ding{51} & \ding{55} & \makecell{\textit{Localization}\\ \textit{Only}} & \textit{Maybe} & $\times 0.5$--$N$ \\
        \bottomrule
    \end{tabular}
    }
\caption{\textbf{A comparison of methods and their requirements.} \textbf{Requires training}: whether the method requires learning additional parameters; \textbf{requires wrong samples}: whether supervision from incorrect examples is needed; \textbf{requires counterfactuals}: whether supervision from counterfactual examples is needed; \textbf{requires decoding}: whether decoding at inference time is needed; \textbf{inference cost}: expected number of forward passes needed, with $N$ decoding steps and $K$ counterfactual samples. \textit{Localization only} indicates the condition is required only for localizing causal variables, but not for correctness prediction; \textit{Maybe} indicates methods that can apply to both prompt tokens and decoded tokens.}
\label{tab:methods}
\end{table}

\begin{table*}[ht]
    \setlength{\tabcolsep}{5pt}
    \centering
    \small
\resizebox{0.75\textwidth}{!}
{
\begin{tabular}{@{}lllccccc@{}}
\toprule
\multicolumn{3}{l}{}                                                             & \textbf{PriceTag}              & \textbf{IOI}                            & \textbf{RAVEL}                          & \textbf{MMLU}                  & \textbf{UnlearnHP}                      \\ \midrule
\multicolumn{8}{l}{\textbf{Output Probabilities}}                                                                                                                                                                                                                                \\
\multicolumn{2}{l}{\multirow{2}{*}{Confidence Score (First N Token)}} & $T$=1    & $0.793_{\hspace{1pt}\pm0.011}$ & $0.935_{\hspace{1pt}\pm0.004}$          & $0.796_{\hspace{1pt}\pm0.023}$          & $\mathbf{0.810}_{\hspace{1pt}\pm0.009}$ & $0.682_{\hspace{1pt}\pm0.016}$          \\
\multicolumn{2}{l}{}                                                  & $T$=2    & $0.804_{\hspace{1pt}\pm0.010}$ & $0.900_{\hspace{1pt}\pm0.016}$          & $0.836_{\hspace{1pt}\pm0.011}$          & $0.795_{\hspace{1pt}\pm0.002}$ & $0.668_{\hspace{1pt}\pm0.023}$          \\
\multicolumn{2}{l}{\multirow{2}{*}{Confidence Score (Answer Token)}}  & $T$=1    & $0.498_{\hspace{1pt}\pm0.002}$ & $0.928_{\hspace{1pt}\pm0.006}$          & $0.795_{\hspace{1pt}\pm0.018}$          & $\mathbf{0.810}_{\hspace{1pt}\pm0.009}$ & $0.664_{\hspace{1pt}\pm0.015}$          \\
\multicolumn{2}{l}{}                                                  & $T$=2    & $0.498_{\hspace{1pt}\pm0.002}$ & $0.870_{\hspace{1pt}\pm0.016}$          & $0.679_{\hspace{1pt}\pm0.015}$          & $0.795_{\hspace{1pt}\pm0.002}$ & $0.657_{\hspace{1pt}\pm0.016}$          \\
\multicolumn{8}{l}{\textbf{Internal Causal-Agnostic Features}}                                                                                                                                                                                                     \\
\multicolumn{3}{l}{Correctness Probing (Background Variables)}    & $0.595_{\hspace{1pt}\pm0.005}$  &  $0.737_{\hspace{1pt}\pm0.021}$  &  $0.859_{\hspace{1pt}\pm0.004}$  &  $0.762_{\hspace{1pt}\pm0.011}$  &  $0.721_{\hspace{1pt}\pm0.025}$ \\
\multicolumn{3}{l}{Correctness Probing (Prompt Last Token)} &   $\mathbf{0.955}_{\hspace{1pt}\pm0.006}$  &  $0.822_{\hspace{1pt}\pm0.023}$  &  $0.895_{\hspace{1pt}\pm0.002}$  &  $\underline{0.804}_{\hspace{1pt}\pm0.007}$  &  $\mathbf{0.940}_{\hspace{1pt}\pm0.009}$ \\

\midrule
\multicolumn{8}{l}{\textbf{Internal Causal Features}}                                                                                                                                                                                                               \\
\multicolumn{3}{l}{Counterfactual Simulation (First N Token)}                     & $0.790_{\hspace{1pt}\pm0.042}$ & $\mathbf{0.999}_{\hspace{1pt}\pm0.000}$ & $\mathbf{0.924}_{\hspace{1pt}\pm0.007}$ & $0.785_{\hspace{1pt}\pm0.009}$ & $0.727_{\hspace{1pt}\pm0.023}$\\
\multicolumn{3}{l}{Counterfactual Simulation (Output Mapping)}                    & $0.907_{\hspace{1pt}\pm0.002}$ & $\mathbf{0.999}_{\hspace{1pt}\pm0.000}$ & $\underline{0.908}_{\hspace{1pt}\pm0.002}$          & $0.785_{\hspace{1pt}\pm0.009}$ & $0.919_{\hspace{1pt}\pm0.010}$          \\
\multicolumn{3}{l}{Value Probing}    & $0.888_{\hspace{1pt}\pm0.021}$  &  $0.804_{\hspace{1pt}\pm0.021}$  &  $0.603_{\hspace{1pt}\pm0.001}$  &  $\underline{0.798}_{\hspace{1pt}\pm0.011}$  &  $0.774_{\hspace{1pt}\pm0.045}$ \\ 
\multicolumn{3}{l}{Correctness Probing (Causal Variables)}      & $\mathbf{0.955}_{\hspace{1pt}\pm0.007}$  &  $0.914_{\hspace{1pt}\pm0.007}$  &  $0.895_{\hspace{1pt}\pm0.002}$  &  $\underline{0.804}_{\hspace{1pt}\pm0.007}$  &  $\mathbf{0.940}_{\hspace{1pt}\pm0.009}$ \\
\bottomrule
\end{tabular}
}
\caption{\textbf{Results of the in-distribution setting.} Overall, methods using causal features are the best. Methods using features from the prompt last token might also have high AUC-ROC if it coincides with the location of causal variables. Numbers are reported in mean$\pm$std. For each task, the highest AUC-ROC value is in \textbf{bold}. Values that are within a standard deviation are \underline{underlined}.}
    \label{tab:main_results}
\end{table*}

\begin{table*}[ht]
    \setlength{\tabcolsep}{3pt}
    \centering
    \small
\resizebox{\textwidth}{!}
{
\begin{tabular}{@{}llcccccccccc@{}}
\toprule
\multicolumn{2}{l}{} &
  \multicolumn{2}{c}{\textbf{PriceTag}} &
  \multicolumn{2}{c}{\textbf{IOI}} &
  \multicolumn{2}{c}{\textbf{RAVEL}} &
  \multicolumn{2}{c}{\textbf{MMLU}} &
  \multicolumn{2}{c}{\textbf{UnlearnHP}} \\ \midrule
\multicolumn{2}{l}{Ways of OOD} &
  \begin{tabular}[c]{@{}c@{}}Change \\ Currency Format\end{tabular} &
  \begin{tabular}[c]{@{}c@{}}Add \\ Typos\end{tabular} &
  \begin{tabular}[c]{@{}c@{}}Change \\ Language\end{tabular} &
  \begin{tabular}[c]{@{}c@{}}Rephrase \\ Templates\end{tabular} &
  \begin{tabular}[c]{@{}c@{}}Change \\ Language\end{tabular} &
  \begin{tabular}[c]{@{}c@{}}Change \\ ICL Demos\end{tabular} &
  \begin{tabular}[c]{@{}c@{}}Change \\ Option Symbols\end{tabular} &
  \begin{tabular}[c]{@{}c@{}}Add \\ Distractor\end{tabular} &
  \begin{tabular}[c]{@{}c@{}}Increase \\ Length\end{tabular} &
  \begin{tabular}[c]{@{}c@{}}Rephrase \\ Template\end{tabular} \\ \midrule
\multicolumn{2}{l}{Task Accuracy} &
  \multirow{2}{*}{96 → 91} &
  \multirow{2}{*}{96 → 80} &
  \multirow{2}{*}{98 → 83} &
  \multirow{2}{*}{98 → 71} &
  \multirow{2}{*}{90 → 86} &
  \multirow{2}{*}{90 → 78} &
  \multirow{2}{*}{65 → 60} &
  \multirow{2}{*}{65 → 55} &
  \multirow{2}{*}{97 → 96} &
  \multirow{2}{*}{97 → 92} \\
\multicolumn{2}{l}{ID → OOD (\%)} &
   &
   &
   &
   &
   &
   &
   &
   &
   &
   \\ \midrule
\multicolumn{8}{l}{\textbf{Output Probabilities}} &
   \\
\multicolumn{8}{l}{Confidence Score} &
   \\
\hspace{1.4em}(First N Token)  &
  $T$=1 &
  $0.631_{\hspace{1pt}\pm0.020}$ &
  $0.777_{\hspace{1pt}\pm0.009}$ &
  $0.767_{\hspace{1pt}\pm0.012}$ &
  $0.650_{\hspace{1pt}\pm0.002}$ &
  $0.874_{\hspace{1pt}\pm0.012}$ &
  $0.693_{\hspace{1pt}\pm0.012}$ &
  $0.707_{\hspace{1pt}\pm0.021}$ &
  $0.770_{\hspace{1pt}\pm0.028}$ &
  $0.626_{\hspace{1pt}\pm0.000}$ &
  $\underline{0.739}_{\hspace{1pt}\pm0.000}$ \\
&
  $T$=2 &
  $0.498_{\hspace{1pt}\pm0.003}$ &
  $0.714_{\hspace{1pt}\pm0.011}$ &
  $0.752_{\hspace{1pt}\pm0.005}$ &
  $0.558_{\hspace{1pt}\pm0.009}$ &
  $0.873_{\hspace{1pt}\pm0.013}$ &
  $0.651_{\hspace{1pt}\pm0.013}$ &
  $\underline{0.761}_{\hspace{1pt}\pm0.017}$ &
  $0.679_{\hspace{1pt}\pm0.009}$ &
  $0.592_{\hspace{1pt}\pm0.000}$ &
  $\underline{0.730}_{\hspace{1pt}\pm0.000}$ \\
\hspace{1.4em}(Answer Token) &
  $T$=1 &
  $0.544_{\hspace{1pt}\pm0.012}$ &
  $0.771_{\hspace{1pt}\pm0.020}$ &
  $0.709_{\hspace{1pt}\pm0.017}$ &
  $0.596_{\hspace{1pt}\pm0.007}$ &
  $\underline{0.908}_{\hspace{1pt}\pm0.016}$ &
  $0.739_{\hspace{1pt}\pm0.011}$ &
  $0.707_{\hspace{1pt}\pm0.021}$ &
  $0.770_{\hspace{1pt}\pm0.028}$ &
  $\mathbf{0.931}_{\hspace{1pt}\pm0.018}$ &
  $0.491_{\hspace{1pt}\pm0.018}$ \\
 &
  $T$=2 &
  $0.498_{\hspace{1pt}\pm0.003}$ &
  $0.729_{\hspace{1pt}\pm0.016}$ &
  $0.671_{\hspace{1pt}\pm0.024}$ &
  $0.521_{\hspace{1pt}\pm0.004}$ &
  $\underline{0.906}_{\hspace{1pt}\pm0.014}$ &
  $0.504_{\hspace{1pt}\pm0.014}$ &
  $\underline{0.761}_{\hspace{1pt}\pm0.017}$ &
  $0.679_{\hspace{1pt}\pm0.009}$ &
  $\underline{0.902}_{\hspace{1pt}\pm0.018}$ &
  $0.484_{\hspace{1pt}\pm0.016}$ \\
\multicolumn{8}{l}{\textbf{Internal Causal-Agnostic Features}} &
   &
   &
   &
   \\
\multicolumn{8}{l}{{Correctness Probing}} &
   &
   &
   &
   \\
\multicolumn{2}{l}{\hspace{1.4em}(Background Variables)} &
  $0.556_{\hspace{1pt}\pm0.003}$ & $0.597_{\hspace{1pt}\pm0.008}$ & $0.541_{\hspace{1pt}\pm0.006}$ & $0.749_{\hspace{1pt}\pm0.017}$ & $0.812_{\hspace{1pt}\pm0.057}$ & $0.837_{\hspace{1pt}\pm0.007}$ & $0.729_{\hspace{1pt}\pm0.013}$ & $0.698_{\hspace{1pt}\pm0.020}$ & $0.656_{\hspace{1pt}\pm0.009}$ & $0.660_{\hspace{1pt}\pm0.009}$ \\

\multicolumn{2}{l}{\hspace{1.4em}(Prompt Last Token)} &
  $0.627_{\hspace{1pt}\pm0.014}$ & $\underline{0.840}_{\hspace{1pt}\pm0.017}$ & $0.607_{\hspace{1pt}\pm0.034}$ & $0.444_{\hspace{1pt}\pm0.073}$ & $0.808_{\hspace{1pt}\pm0.033}$ & $0.858_{\hspace{1pt}\pm0.019}$ & $\mathbf{0.784}_{\hspace{1pt}\pm0.006}$ & $0.694_{\hspace{1pt}\pm0.012}$ & $0.830_{\hspace{1pt}\pm0.022}$ & $0.648_{\hspace{1pt}\pm0.005}$ \\

\midrule
\multicolumn{8}{l}{\textbf{Internal Causal Features}} &
   &
   &
   &
   \\
\multicolumn{12}{l}{Counterfactual Simulation} \\
\multicolumn{2}{l}{\hspace{1.4em}(First N Token)} &
  $0.624_{\hspace{1pt}\pm0.066}$ & $\underline{0.872}_{\hspace{1pt}\pm0.051}$ & $\underline{0.996}_{\hspace{1pt}\pm0.002}$ & $\mathbf{0.816}_{\hspace{1pt}\pm0.014}$ & $\underline{0.923}_{\hspace{1pt}\pm0.026}$ & $\underline{0.918}_{\hspace{1pt}\pm0.042}$ & $0.765_{\hspace{1pt}\pm0.002}$ & $\mathbf{0.810}_{\hspace{1pt}\pm0.011}$ & $0.725_{\hspace{1pt}\pm0.035}$ & $0.655_{\hspace{1pt}\pm0.020}$ \\
\multicolumn{2}{l}{\hspace{1.4em}(Output Mapping)} &
  $\mathbf{0.856}_{\hspace{1pt}\pm0.024}$ &
  $\mathbf{0.875}_{\hspace{1pt}\pm0.050}$ &
  $\mathbf{0.997}_{\hspace{1pt}\pm0.001}$ &
  $\underline{0.815}_{\hspace{1pt}\pm0.004}$ &
  $\mathbf{0.939}_{\hspace{1pt}\pm0.022}$ &
  $\mathbf{0.931}_{\hspace{1pt}\pm0.029}$ &
  $0.765_{\hspace{1pt}\pm0.002}$ &
  $\mathbf{0.810}_{\hspace{1pt}\pm0.011}$ &
  $0.860_{\hspace{1pt}\pm0.011}$ &
  $\mathbf{0.772}_{\hspace{1pt}\pm0.056}$ \\

\multicolumn{2}{l}{Value Probing} &
  $0.777_{\hspace{1pt}\pm0.026}$ & $\underline{0.870}_{\hspace{1pt}\pm0.063}$ & $0.611_{\hspace{1pt}\pm0.063}$ & $\underline{0.805}_{\hspace{1pt}\pm0.010}$ & $0.691_{\hspace{1pt}\pm0.040}$ & $0.652_{\hspace{1pt}\pm0.027}$ & $\underline{0.777}_{\hspace{1pt}\pm0.015}$ & $0.774_{\hspace{1pt}\pm0.007}$ & $0.846_{\hspace{1pt}\pm0.012}$ & $0.649_{\hspace{1pt}\pm0.030}$ \\ 
\multicolumn{8}{l}{{Correctness Probing}} &
&
&
&
\\
  \multicolumn{2}{l}{\hspace{1.4em}(Causal Variables)} &
  $0.693_{\hspace{1pt}\pm0.002}$ & $\underline{0.862}_{\hspace{1pt}\pm0.002}$ & $0.806_{\hspace{1pt}\pm0.023}$ & $0.665_{\hspace{1pt}\pm0.032}$ & $\underline{0.910}_{\hspace{1pt}\pm0.015}$ & $\underline{0.917}_{\hspace{1pt}\pm0.007}$ & $\mathbf{0.784}_{\hspace{1pt}\pm0.006}$ & $0.694_{\hspace{1pt}\pm0.011}$ & $0.828_{\hspace{1pt}\pm0.023}$ & $0.650_{\hspace{1pt}\pm0.006}$ \\
  \bottomrule
\end{tabular}
}
    \vspace{-1ex}
    \caption{\textbf{Results of the out-of-distribution setting.} Methods using internal causal features are significantly more robust than methods using other features. Causal Simulation performs the best on 8 out of the 10 tasks. %
    }
    \label{tab:ood_results}
    \vspace{-1ex}
\end{table*}

We evaluate four correctness prediction methods over a suite of five language modeling tasks under both in-distribution and OOD settings, primarily using the \texttt{Llama-3-8B-Instruct} model~\cite{llama3modelcard}.\footnote{Data and code available at \url{https://github.com/explanare/ood-prediction}}

\subsection{Setup}
\label{sec:exp_setup}

\paragraph{Evaluation tasks} We consider a variety of tasks that cover both idealized cases where we know the task mechanisms and open-ended ones where
only partial or approximate mechanisms are identified. These tasks fall into three categories: (1) \textit{symbol manipulation tasks}, including Indirect Object Identification (IOI; \citealp{wang2023ioi}) and PriceTag \cite{wu2023interpretabilityatscale}, for which the internal mechanisms used to solve the task are clearly known; (2) \textit{knowledge retrieval tasks}, including RAVEL \cite{huang-etal-2024-ravel} and MMLU \cite{hendrycks2021measuringmassivemultitasklanguage}, whose mechanisms are only partially understood; (3) an \textit{instruction following task}: UnlearnHP \cite{thaker2024guardrailbaselinesunlearningllms}, whose mechanisms are similarly only partially known.
These tasks include both constrained and open-ended output formats. For each task, we select prompts on which the model achieves relatively high accuracy in behavioral testing. Further details on tasks and prompts are provided in Appendix~\ref{appx:dataset}.

\paragraph{Metrics}
We construct a test set with an equal number of correctly and wrongly predicted inputs, which turns the correctness prediction task into a balanced binary classification task. Following the evaluation setup of similar truthfulness prediction tasks~\cite{orgad2024llmsknowshowintrinsic,yin2024truthfulness}, we measure the prediction accuracy using AUC-ROC.

\paragraph{ID vs.~OOD setting} For each task, we consider two evaluation settings. In the ID setting, both the test set and the verified set are randomly sampled; in the OOD setting, we introduce perturbations to the inputs in the test split, changing only task-irrelevant background conditions while preserving the underlying task structure. We use perturbations commonly employed for stress testing language models, such as paraphrasing templates, translating to a different language, changing in-context demos, and adding spelling variations and distracting sequences. We specifically select perturbations that lead to a notable drop in model accuracy. See details on OOD prompt selection in Section~\ref{appx:prompt_selection}.
\begin{figure*}[ht]
    \centering
   \resizebox{\textwidth}{!}{
    \includegraphics{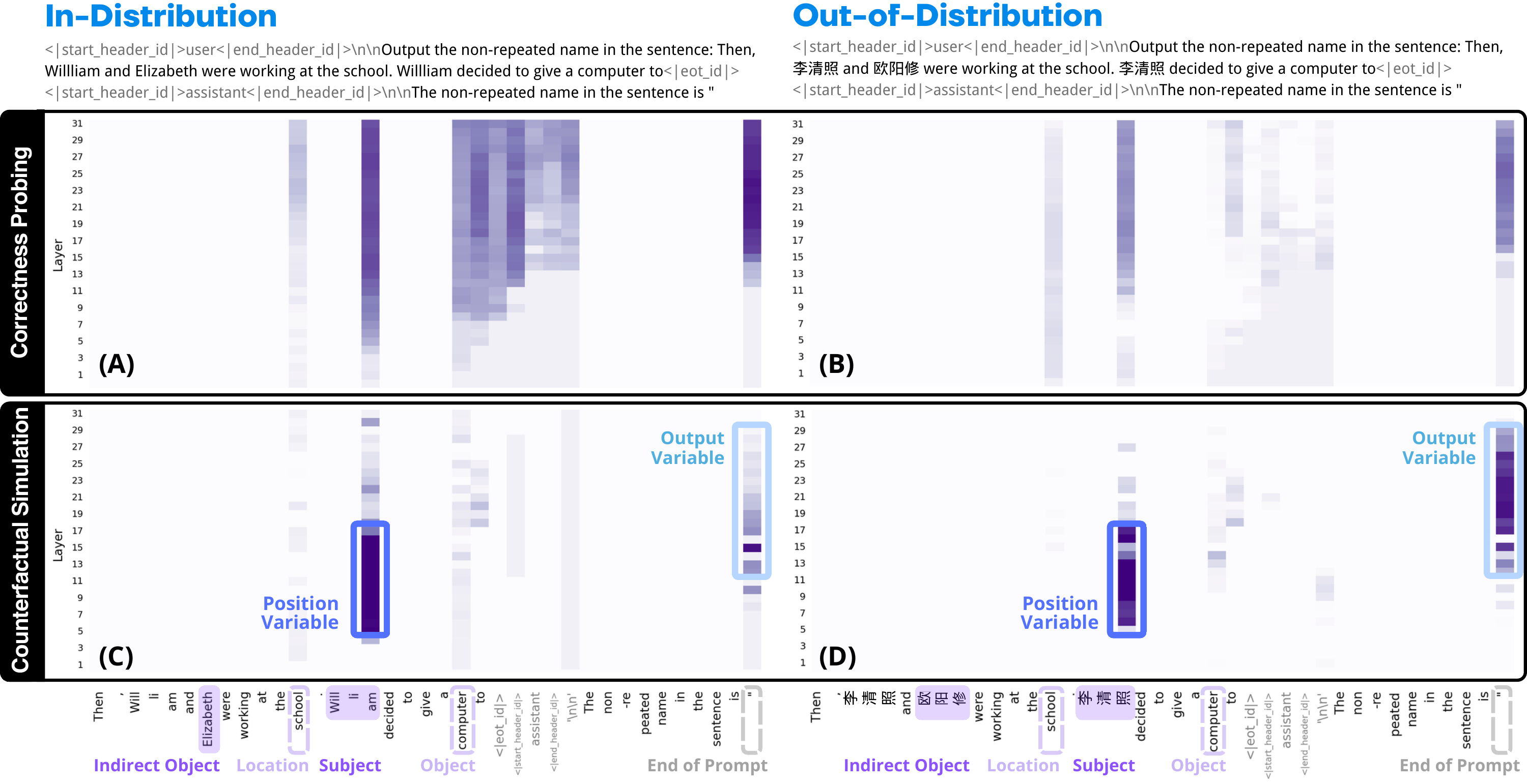}
    }
    \vspace{-3ex}
    \caption{\textbf{Robust vs. non-robust features.} Many features that are predictive of correctness on in-distribution inputs (purple blocks in panel A), where only very few of them can generalize to out-of-distribution inputs (\textbf{\textcolor{RoyalPurple}{purple blocks}} in panel B). These features coincide with features that have causal effects on task predictions, as shown in panel C and D.}
    \vspace{-1ex}
    \label{fig:aucroc_vs_loc}
\end{figure*}

\paragraph{Features and methods}
We evaluate different combinations of features and methods. For features, we consider output probabilities (Section~\ref{sec:output_probabilities}), internal causal-agnostic features (Section~\ref{sec:causal_agnostic}), and internal causal features (Section~\ref{sec:causal}). For causal-agnostic features, we further categorize them based on the feature locations, i.e., ``prompt last token'', and the actual causal roles these features played, i.e., ``background variable'' vs.\ ``causal variable''.
All internal features are extracted from residual streams, since residual representations serve as an information bottleneck in Transformers.

Orthogonal to the choice of features, we evaluate four methods for correctness prediction (summarized in Table~\ref{tab:methods}). These methods can be applied to different features (e.g., correctness probing can be applied to causal variables as well, as shown in the last row of the table). Each method is associated with a different set of hyperparameters, whose optimal values are selected using a validation set drawn from the training distribution.

\subsection{Results}

\paragraph{In-distribution results}
Table~\ref{tab:main_results} presents the in-distribution results. Comparing across the four feature groups, we observe that features extracted from causal variables are the most predictive of correctness, followed by features from the last prompt token. 
Comparing across methods, counterfactual simulation achieves the best results, especially on IOI and RAVEL. Correctness probing based on causal features performs comparably or slightly better than counterfactual simulation (by approximately 1--2\%), especially on knowledge retrieval tasks where internal mechanisms are not fully known (MMLU, UnlearnHP). Value probing generally performs worse than correctness probing and counterfactual simulation.
Confidence scores (output probabilities) are also very predictive of correctness when task outputs are highly constrained (IOI, MMLU, and RAVEL).
Further experiments with Llama 3B, 13B, and 70B models on a factual retrieval task, i.e., RAVEL, indicate that the predictive power of confidence scores does not increase much as the model scales up (Appendix~\ref{appx:model_variation_results}).

\paragraph{Out-of-distribution results}
 Table~\ref{tab:ood_results} summarizes the out-of-distribution results. Comparing feature locations, methods based on internal causal features \textit{significantly} outperform those relying on non-causal features.
Comparing across methods, counterfactual simulation achieves the strongest overall performance, outperforming other methods on 8 out of 10 OOD tasks. Correctness probing based on causal variables also performs robustly but typically ranks behind counterfactual simulation. Value probing achieves competitive results only occasionally, generally performing worse than other causal methods.
In contrast, confidence scores become significantly less predictive under distribution shifts. Moreover, their predictive power is sensitive to the choice of temperature, and the optimal temperature does not generalize well across in-distribution and OOD settings. An additional experiment (Appendix~\ref{appx:ravel_ood_two_shot}) further demonstrates that confidence scores become notably less predictive as task accuracy declines, whereas counterfactual simulation maintains stable performance.

\section{Discussion}
\subsection{Which Features Are the Most Robust?}

Our results in Table~\ref{tab:ood_results} show that causal features are the more robust features for correctness prediction compared with others. This is further demonstrated in Figure~\ref{fig:aucroc_vs_loc}. This finding aligns precisely with the intuition motivating this work: only internal features that are \textit{used} by the model to make predictions can reliably signal model behavior---that is, they must causally contribute to the model's success at solving the task. Moreover, features used \textit{only} for specific inputs should not remain reliable under distribution shifts.

Comparing across feature locations, we find that features extracted from the last prompt token are also predictive on some prompts, when they are on a direct causal path from causal variables to outputs, effectively mediating the causal influence on outputs.
They may also play a direct causal role when the last prompt token coincides with the answer token. Needless to say, we cannot rely on this particular pattern in most settings.

\subsection{When to Use Which Methods?}
Unlike features, there is no single method that proves optimal under all settings. We observe at least two factors that affect correctness prediction performance.

\paragraph{Task properties}
The choice of methods depends crucially on the properties of the task. Three task properties matter: (1) whether the task involves symbolic computation or merely memorization of training data; (2) how much we already know about the mechanisms shared across task examples; and (3) whether the task output is free-formed or constrained. In practice, these properties often co-occur (e.g., knowledge retrieval tasks usually require memorization and constrained outputs containing specific proper nouns).

Considering (1),  symbolic tasks (e.g., IOI, PriceTag), whose task mechanisms are readily cashed out using causal models, strongly favor causal feature-based methods, with counterfactual simulation performing best. 
Tasks primarily involving memorization (e.g., RAVEL, MMLU) favor confidence scores, followed by correctness probing.
Considering (2), causal methods do not outperform confidence scores on MMLU, likely because it covers mixed subjects (e.g., international law, abstract algebra) sharing only a multiple-choice structure, with the consequence that we know only a very incomplete high-level model. %
Finally, considering (3), confidence scores perform well on tasks with predefined, constrained outputs (e.g., multiple-choice questions like MMLU), but struggle with open-ended outputs (e.g., PriceTag). This limitation arises from the lack of  principled methods for assigning exact probabilities to individual events in free-form text outputs~\cite{kuleshov2015calibrated}.

\paragraph{Practical considerations}
As shown in Table~\ref{tab:methods}, these methods also differ substantially in the types of training data, the amount of prior knowledge, and the inference-time computation required. For tasks where incorrect examples are rare and expensive to collect, methods that do not require negative examples have a clear advantage. For settings where the computation cost is critical, probing methods that avoid decoding can be ideal.

\subsection{Is Simulating Counterfactual Behaviors Aligned with Predicting Model Behaviors?}

In this section, we examine whether advancements in the \textit{abstraction} direction (i.e., developing better methods for finding internal causal mechanisms) align with the advancements in the \textit{prediction} direction (i.e., more accurately predicting the model's output correctness).

\begin{figure}[t]
    \centering
   \resizebox{1\columnwidth}{!}{
    \includegraphics{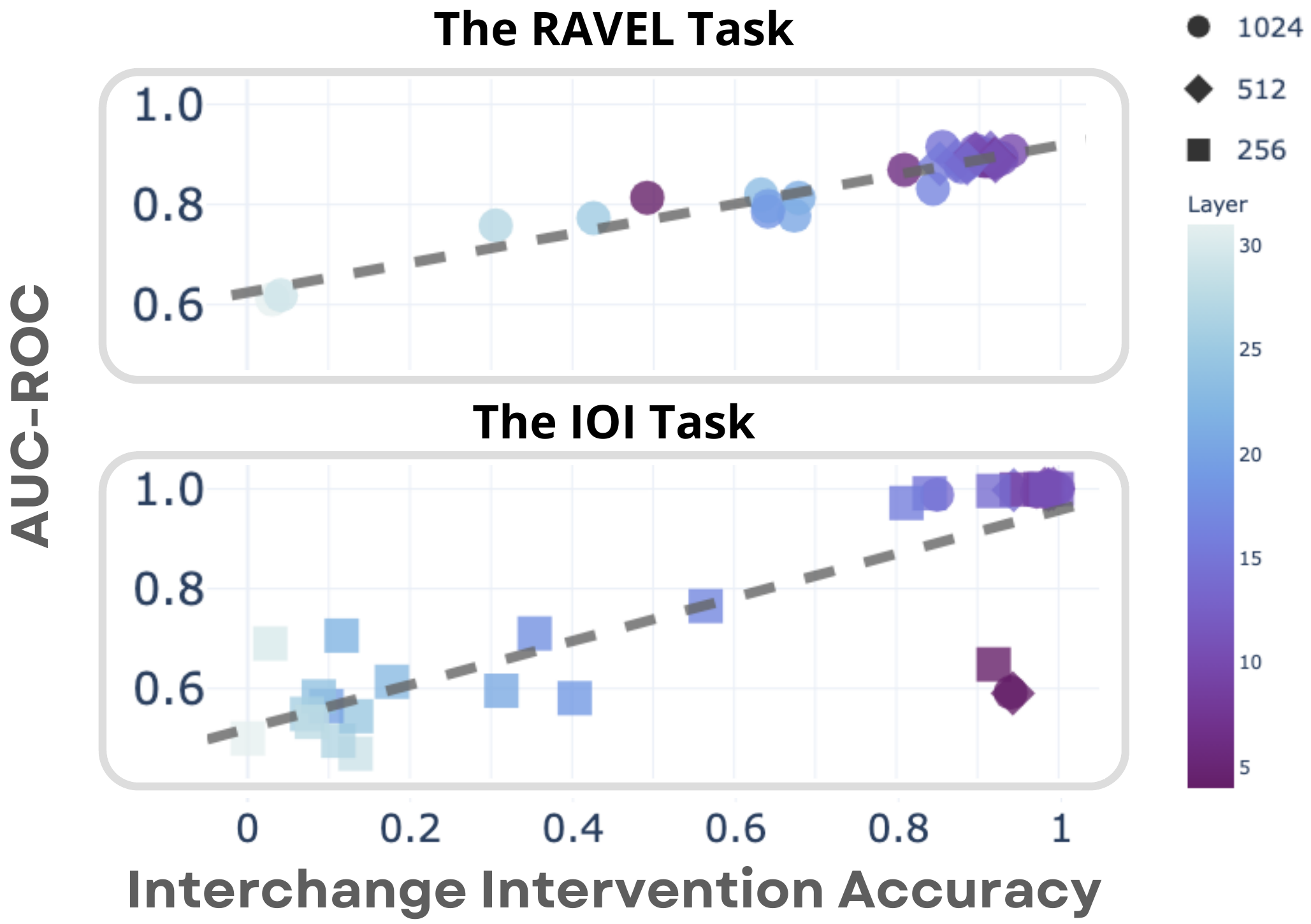}
    }
    \vspace{-2ex}
    \caption{\textbf{A positive correlation between interchange intervention accuracy and AUC-ROC in in-distribution settings.} Interchange intervention accuracy measures the ability to simulate causal effects, and AUC-ROC measures the ability to predict model correctness. Improving the ability to simulate causal effects generally improves the ability to predict the correctness of model outputs. Each point represents one observation. Different shapes represent different dimensions used to train DAS.}
    \label{fig:aucroc_vs_iia}
    \vspace{-1ex}
\end{figure}

\paragraph{Strong correlation between accuracy on simulating counterfactual behaviors and predicting model behavior} 
Figure~\ref{fig:aucroc_vs_iia} shows an empirical investigation of the relationship between IIA and AUC-ROC. We observe a positive correlation: as IIA increases, AUC-ROC also increases, meaning the localized causal features become more predictive of correctness. Though once IIA approaches 1, this correlation no longer holds; further increases in IIA no longer yield improvements in AUC-ROC.
Data points at the lower right of Figure~\ref{fig:aucroc_vs_iia} indicate that certain representations in earlier layers may have strong causal effects (as suggested by high IIA scores) but remain weak predictors of OOD correctness. This likely occurs because these representations are easily computed by the model regardless of whether the final prediction is correct or incorrect.

\paragraph{Correctness estimation as an additional evaluation objective for finding useful causal mechanisms} %
Given two representations that localize a high-level variable with similar IIA, we might prefer the representation with high AUC-ROC, which would help us better predict model behaviors.

\section{Conclusions}
We have shown that internal causal features are the most robust features for predicting output correctness, especially under distribution shifts. We propose two methods---counterfactual simulation and value probing---that outperform causal-agnostic approaches in out-of-distribution settings. More broadly, we introduce a novel framework that draws upon causal analysis of the model's internal mechanisms to predict its generalization behaviors. This suggests a promising path forward: internal causal analysis of language models is not only explanatory, but also predictive, contributing to reliability and safety in practical deployments of language models.

\section*{Acknowledgments}
We are grateful to the anonymous reviewers for their valuable feedback. We would also like to thank Atticus Geiger, Xiaoyin Chen, and Zhengxuan Wu for their insightful comments and helpful discussions. This research is supported in part by grants from Google, Open Philanthropy, ONR grant N000142412532, and NSF grant IIS-2247357.

\section*{Impact Statement} 

This paper presents work whose goal is to advance the field of Machine Learning. There are many potential societal consequences of our work, none of which we feel must be specifically highlighted here.

\bibliography{anthology,icml2024}

\begin{thebibliography}{51}
\providecommand{\natexlab}[1]{#1}
\providecommand{\url}[1]{\texttt{#1}}
\expandafter\ifx\csname urlstyle\endcsname\relax
  \providecommand{\doi}[1]{doi: #1}\else
  \providecommand{\doi}{doi: \begingroup \urlstyle{rm}\Url}\fi

\bibitem[AI@Meta(2024)]{llama3modelcard}
AI@Meta.
\newblock Llama 3 model card, 2024.
\newblock URL \url{https://github.com/meta-llama/llama3/blob/main/MODEL_CARD.md}.

\bibitem[Arditi et~al.(2024)Arditi, Obeso, Syed, Paleka, Panickssery, Gurnee, and Nanda]{arditi2024refusal}
Arditi, A., Obeso, O., Syed, A., Paleka, D., Panickssery, N., Gurnee, W., and Nanda, N.
\newblock Refusal in language models is mediated by a single direction.
\newblock In Globerson, A., Mackey, L., Belgrave, D., Fan, A., Paquet, U., Tomczak, J., and Zhang, C. (eds.), \emph{Advances in Neural Information Processing Systems}, volume~37, pp.\  136037--136083. Curran Associates, Inc., 2024.
\newblock URL \url{https://proceedings.neurips.cc/paper_files/paper/2024/file/f545448535dfde4f9786555403ab7c49-Paper-Conference.pdf}.

\bibitem[Arora et~al.(2024)Arora, Jurafsky, and Potts]{arora-etal-2024-causalgym}
Arora, A., Jurafsky, D., and Potts, C.
\newblock {C}ausal{G}ym: Benchmarking causal interpretability methods on linguistic tasks.
\newblock In Ku, L.-W., Martins, A., and Srikumar, V. (eds.), \emph{Proceedings of the 62nd Annual Meeting of the Association for Computational Linguistics (Volume 1: Long Papers)}, pp.\  14638--14663, Bangkok, Thailand, August 2024. Association for Computational Linguistics.
\newblock \doi{10.18653/v1/2024.acl-long.785}.
\newblock URL \url{https://aclanthology.org/2024.acl-long.785}.

\bibitem[Azaria \& Mitchell(2023)Azaria and Mitchell]{azaria-mitchell-2023-internal}
Azaria, A. and Mitchell, T.
\newblock The internal state of an {LLM} knows when it{'}s lying.
\newblock In Bouamor, H., Pino, J., and Bali, K. (eds.), \emph{Findings of the Association for Computational Linguistics: EMNLP 2023}, pp.\  967--976, Singapore, December 2023. Association for Computational Linguistics.
\newblock \doi{10.18653/v1/2023.findings-emnlp.68}.
\newblock URL \url{https://aclanthology.org/2023.findings-emnlp.68}.

\bibitem[Bills et~al.(2023)Bills, Cammarata, Mossing, Tillman, Gao, Goh, Sutskever, Leike, Wu, and Saunders]{bills2023language}
Bills, S., Cammarata, N., Mossing, D., Tillman, H., Gao, L., Goh, G., Sutskever, I., Leike, J., Wu, J., and Saunders, W.
\newblock Language models can explain neurons in language models.
\newblock \emph{URL https://openaipublic. blob. core. windows. net/neuron-explainer/paper/index. html.(Date accessed: 14.05. 2023)}, 2, 2023.

\bibitem[Chuang et~al.(2024)Chuang, Qiu, Hsieh, Krishna, Kim, and Glass]{chuang-etal-2024-lookback}
Chuang, Y.-S., Qiu, L., Hsieh, C.-Y., Krishna, R., Kim, Y., and Glass, J.~R.
\newblock Lookback lens: Detecting and mitigating contextual hallucinations in large language models using only attention maps.
\newblock In Al-Onaizan, Y., Bansal, M., and Chen, Y.-N. (eds.), \emph{Proceedings of the 2024 Conference on Empirical Methods in Natural Language Processing}, pp.\  1419--1436, Miami, Florida, USA, November 2024. Association for Computational Linguistics.
\newblock \doi{10.18653/v1/2024.emnlp-main.84}.
\newblock URL \url{https://aclanthology.org/2024.emnlp-main.84}.

\bibitem[Conmy et~al.(2023)Conmy, Mavor{-}Parker, Lynch, Heimersheim, and Garriga{-}Alonso]{conmy2023acdc}
Conmy, A., Mavor{-}Parker, A.~N., Lynch, A., Heimersheim, S., and Garriga{-}Alonso, A.
\newblock Towards automated circuit discovery for mechanistic interpretability.
\newblock In Oh, A., Naumann, T., Globerson, A., Saenko, K., Hardt, M., and Levine, S. (eds.), \emph{Advances in Neural Information Processing Systems 36: Annual Conference on Neural Information Processing Systems 2023, NeurIPS 2023, New Orleans, LA, USA, December 10 - 16, 2023}, 2023.

\bibitem[Desai \& Durrett(2020)Desai and Durrett]{desai-durrett-2020-calibration}
Desai, S. and Durrett, G.
\newblock Calibration of pre-trained transformers.
\newblock In Webber, B., Cohn, T., He, Y., and Liu, Y. (eds.), \emph{Proceedings of the 2020 Conference on Empirical Methods in Natural Language Processing (EMNLP)}, pp.\  295--302, Online, November 2020. Association for Computational Linguistics.
\newblock \doi{10.18653/v1/2020.emnlp-main.21}.
\newblock URL \url{https://aclanthology.org/2020.emnlp-main.21}.

\bibitem[Du et~al.(2024)Du, Xiao, and Li]{du2024haloscope}
Du, X., Xiao, C., and Li, Y.
\newblock Haloscope: Harnessing unlabeled {LLM} generations for hallucination detection.
\newblock In \emph{The Thirty-eighth Annual Conference on Neural Information Processing Systems}, 2024.
\newblock URL \url{https://openreview.net/forum?id=nfK0ZXFFSn}.

\bibitem[Engels et~al.(2025)Engels, Michaud, Liao, Gurnee, and Tegmark]{engels2025not}
Engels, J., Michaud, E.~J., Liao, I., Gurnee, W., and Tegmark, M.
\newblock Not all language model features are one-dimensionally linear.
\newblock In \emph{The Thirteenth International Conference on Learning Representations}, 2025.
\newblock URL \url{https://openreview.net/forum?id=d63a4AM4hb}.

\bibitem[Ferrando et~al.(2025)Ferrando, Obeso, Rajamanoharan, and Nanda]{ferrando2024knowledgeawareness}
Ferrando, J., Obeso, O., Rajamanoharan, S., and Nanda, N.
\newblock Do i know this entity? knowledge awareness and hallucinations in language models.
\newblock In \emph{International Conference on Learning Representations}, 2025.
\newblock URL \url{https://openreview.net/forum?id=WCRQFlji2q}.

\bibitem[Finlayson et~al.(2021)Finlayson, Mueller, Gehrmann, Shieber, Linzen, and Belinkov]{finlayson-etal-2021-causal}
Finlayson, M., Mueller, A., Gehrmann, S., Shieber, S., Linzen, T., and Belinkov, Y.
\newblock Causal analysis of syntactic agreement mechanisms in neural language models.
\newblock In Zong, C., Xia, F., Li, W., and Navigli, R. (eds.), \emph{Proceedings of the 59th Annual Meeting of the Association for Computational Linguistics and the 11th International Joint Conference on Natural Language Processing (Volume 1: Long Papers)}, pp.\  1828--1843, Online, August 2021. Association for Computational Linguistics.
\newblock \doi{10.18653/v1/2021.acl-long.144}.
\newblock URL \url{https://aclanthology.org/2021.acl-long.144}.

\bibitem[Geiger et~al.(2021)Geiger, Lu, Icard, and Potts]{geiger2021causal}
Geiger, A., Lu, H., Icard, T., and Potts, C.
\newblock Causal abstractions of neural networks.
\newblock In \emph{Advances in Neural Information Processing Systems}, volume~34, pp.\  9574--9586, 2021.
\newblock URL \url{https://papers.nips.cc/paper/2021/hash/4f5c422f4d49a5a807eda27434231040-Abstract.html}.

\bibitem[Geiger et~al.(2024{\natexlab{a}})Geiger, Ibeling, Zur, Chaudhary, Chauhan, Huang, Arora, Wu, Goodman, Potts, and Icard]{geiger2024causalabstraction}
Geiger, A., Ibeling, D., Zur, A., Chaudhary, M., Chauhan, S., Huang, J., Arora, A., Wu, Z., Goodman, N., Potts, C., and Icard, T.
\newblock Causal abstraction: A theoretical foundation for mechanistic interpretability, 2024{\natexlab{a}}.
\newblock URL \url{https://arxiv.org/abs/2301.04709}.

\bibitem[Geiger et~al.(2024{\natexlab{b}})Geiger, Wu, Potts, Icard, and Goodman]{geiger2024das}
Geiger, A., Wu, Z., Potts, C., Icard, T., and Goodman, N.
\newblock Finding alignments between interpretable causal variables and distributed neural representations.
\newblock In Locatello, F. and Didelez, V. (eds.), \emph{Proceedings of the Third Conference on Causal Learning and Reasoning}, volume 236 of \emph{Proceedings of Machine Learning Research}, pp.\  160--187. PMLR, 01--03 Apr 2024{\natexlab{b}}.
\newblock URL \url{https://proceedings.mlr.press/v236/geiger24a.html}.

\bibitem[Gottesman \& Geva(2024)Gottesman and Geva]{gottesman-geva-2024-estimating}
Gottesman, D. and Geva, M.
\newblock Estimating knowledge in large language models without generating a single token.
\newblock In Al-Onaizan, Y., Bansal, M., and Chen, Y.-N. (eds.), \emph{Proceedings of the 2024 Conference on Empirical Methods in Natural Language Processing}, pp.\  3994--4019, Miami, Florida, USA, November 2024. Association for Computational Linguistics.
\newblock \doi{10.18653/v1/2024.emnlp-main.232}.
\newblock URL \url{https://aclanthology.org/2024.emnlp-main.232}.

\bibitem[Guo et~al.(2017)Guo, Pleiss, Sun, and Weinberger]{guo2017calibration}
Guo, C., Pleiss, G., Sun, Y., and Weinberger, K.~Q.
\newblock On calibration of modern neural networks.
\newblock In Precup, D. and Teh, Y.~W. (eds.), \emph{Proceedings of the 34th International Conference on Machine Learning}, volume~70 of \emph{Proceedings of Machine Learning Research}, pp.\  1321--1330. PMLR, 06--11 Aug 2017.
\newblock URL \url{https://proceedings.mlr.press/v70/guo17a.html}.

\bibitem[Hendrycks \& Gimpel(2017)Hendrycks and Gimpel]{hendrycks2017a}
Hendrycks, D. and Gimpel, K.
\newblock A baseline for detecting misclassified and out-of-distribution examples in neural networks.
\newblock In \emph{International Conference on Learning Representations}, 2017.
\newblock URL \url{https://openreview.net/forum?id=Hkg4TI9xl}.

\bibitem[Hendrycks et~al.(2021)Hendrycks, Burns, Basart, Zou, Mazeika, Song, and Steinhardt]{hendrycks2021measuringmassivemultitasklanguage}
Hendrycks, D., Burns, C., Basart, S., Zou, A., Mazeika, M., Song, D., and Steinhardt, J.
\newblock Measuring massive multitask language understanding.
\newblock In \emph{International Conference on Learning Representations}, 2021.
\newblock URL \url{https://openreview.net/forum?id=d7KBjmI3GmQ}.

\bibitem[Hennigen et~al.(2024)Hennigen, Shen, Nrusimha, Gapp, Sontag, and Kim]{hennigen2024symgen}
Hennigen, L.~T., Shen, S., Nrusimha, A., Gapp, B., Sontag, D., and Kim, Y.
\newblock Towards verifiable text generation with symbolic references.
\newblock In \emph{Proceedings of the Conference on Language Modeling (COLM)}, 2024.
\newblock URL \url{https://arxiv.org/abs/2311.09188}.

\bibitem[Hernandez et~al.(2022)Hernandez, Schwettmann, Bau, Bagashvili, Torralba, and Andreas]{hernandez2022natural}
Hernandez, E., Schwettmann, S., Bau, D., Bagashvili, T., Torralba, A., and Andreas, J.
\newblock Natural language descriptions of deep features.
\newblock In \emph{International Conference on Learning Representations}, 2022.
\newblock URL \url{https://openreview.net/forum?id=NudBMY-tzDr}.

\bibitem[Huang et~al.(2023)Huang, Geiger, D'Oosterlinck, Wu, and Potts]{huang2023rigorously}
Huang, J., Geiger, A., D'Oosterlinck, K., Wu, Z., and Potts, C.
\newblock Rigorously assessing natural language explanations of neurons.
\newblock In Belinkov, Y., Hao, S., Jumelet, J., Kim, N., McCarthy, A., and Mohebbi, H. (eds.), \emph{Proceedings of the 6th BlackboxNLP Workshop: Analyzing and Interpreting Neural Networks for NLP, BlackboxNLP@EMNLP 2023, Singapore, December 7, 2023}, pp.\  317--331. Association for Computational Linguistics, 2023.
\newblock \doi{10.18653/V1/2023.BLACKBOXNLP-1.24}.
\newblock URL \url{https://doi.org/10.18653/v1/2023.blackboxnlp-1.24}.

\bibitem[Huang et~al.(2024)Huang, Wu, Potts, Geva, and Geiger]{huang-etal-2024-ravel}
Huang, J., Wu, Z., Potts, C., Geva, M., and Geiger, A.
\newblock {RAVEL}: Evaluating interpretability methods on disentangling language model representations.
\newblock In Ku, L.-W., Martins, A., and Srikumar, V. (eds.), \emph{Proceedings of the 62nd Annual Meeting of the Association for Computational Linguistics (Volume 1: Long Papers)}, pp.\  8669--8687, Bangkok, Thailand, August 2024. Association for Computational Linguistics.
\newblock \doi{10.18653/v1/2024.acl-long.470}.
\newblock URL \url{https://aclanthology.org/2024.acl-long.470}.

\bibitem[Ji et~al.(2024)Ji, Chen, Ishii, Cahyawijaya, Bang, Wilie, and Fung]{ji-etal-2024-llm}
Ji, Z., Chen, D., Ishii, E., Cahyawijaya, S., Bang, Y., Wilie, B., and Fung, P.
\newblock {LLM} internal states reveal hallucination risk faced with a query.
\newblock In Belinkov, Y., Kim, N., Jumelet, J., Mohebbi, H., Mueller, A., and Chen, H. (eds.), \emph{Proceedings of the 7th BlackboxNLP Workshop: Analyzing and Interpreting Neural Networks for NLP}, pp.\  88--104, Miami, Florida, US, November 2024. Association for Computational Linguistics.
\newblock \doi{10.18653/v1/2024.blackboxnlp-1.6}.
\newblock URL \url{https://aclanthology.org/2024.blackboxnlp-1.6}.

\bibitem[Jia \& Liang(2017)Jia and Liang]{jia-liang-2017-adversarial}
Jia, R. and Liang, P.
\newblock Adversarial examples for evaluating reading comprehension systems.
\newblock In Palmer, M., Hwa, R., and Riedel, S. (eds.), \emph{Proceedings of the 2017 Conference on Empirical Methods in Natural Language Processing}, pp.\  2021--2031, Copenhagen, Denmark, September 2017. Association for Computational Linguistics.
\newblock \doi{10.18653/v1/D17-1215}.
\newblock URL \url{https://aclanthology.org/D17-1215}.

\bibitem[Kadavath et~al.(2022)Kadavath, Conerly, Askell, Henighan, Drain, Perez, Schiefer, Hatfield-Dodds, DasSarma, Tran-Johnson, Johnston, El-Showk, Jones, Elhage, Hume, Chen, Bai, Bowman, Fort, Ganguli, Hernandez, Jacobson, Kernion, Kravec, Lovitt, Ndousse, Olsson, Ringer, Amodei, Brown, Clark, Joseph, Mann, McCandlish, Olah, and Kaplan]{kadavath2022languagemodelsmostlyknow}
Kadavath, S., Conerly, T., Askell, A., Henighan, T., Drain, D., Perez, E., Schiefer, N., Hatfield-Dodds, Z., DasSarma, N., Tran-Johnson, E., Johnston, S., El-Showk, S., Jones, A., Elhage, N., Hume, T., Chen, A., Bai, Y., Bowman, S., Fort, S., Ganguli, D., Hernandez, D., Jacobson, J., Kernion, J., Kravec, S., Lovitt, L., Ndousse, K., Olsson, C., Ringer, S., Amodei, D., Brown, T., Clark, J., Joseph, N., Mann, B., McCandlish, S., Olah, C., and Kaplan, J.
\newblock Language models (mostly) know what they know, 2022.
\newblock URL \url{https://arxiv.org/abs/2207.05221}.

\bibitem[Kuleshov \& Liang(2015)Kuleshov and Liang]{kuleshov2015calibrated}
Kuleshov, V. and Liang, P.~S.
\newblock Calibrated structured prediction.
\newblock In Cortes, C., Lawrence, N., Lee, D., Sugiyama, M., and Garnett, R. (eds.), \emph{Advances in Neural Information Processing Systems}, volume~28. Curran Associates, Inc., 2015.
\newblock URL \url{https://proceedings.neurips.cc/paper_files/paper/2015/file/52d2752b150f9c35ccb6869cbf074e48-Paper.pdf}.

\bibitem[Lakshminarayanan et~al.(2017)Lakshminarayanan, Pritzel, and Blundell]{lakshminarayanan2017ensembles}
Lakshminarayanan, B., Pritzel, A., and Blundell, C.
\newblock Simple and scalable predictive uncertainty estimation using deep ensembles.
\newblock In Guyon, I., Luxburg, U.~V., Bengio, S., Wallach, H., Fergus, R., Vishwanathan, S., and Garnett, R. (eds.), \emph{Advances in Neural Information Processing Systems}, volume~30. Curran Associates, Inc., 2017.
\newblock URL \url{https://proceedings.neurips.cc/paper_files/paper/2017/file/9ef2ed4b7fd2c810847ffa5fa85bce38-Paper.pdf}.

\bibitem[Marks \& Tegmark(2024)Marks and Tegmark]{marks2024the}
Marks, S. and Tegmark, M.
\newblock The geometry of truth: Emergent linear structure in large language model representations of true/false datasets.
\newblock In \emph{First Conference on Language Modeling}, 2024.
\newblock URL \url{https://openreview.net/forum?id=aajyHYjjsk}.

\bibitem[Meng et~al.(2023)Meng, Sharma, Andonian, Belinkov, and Bau]{meng2023massediting}
Meng, K., Sharma, A.~S., Andonian, A.~J., Belinkov, Y., and Bau, D.
\newblock Mass-editing memory in a transformer.
\newblock In \emph{The Eleventh International Conference on Learning Representations}, 2023.
\newblock URL \url{https://openreview.net/forum?id=MkbcAHIYgyS}.

\bibitem[Mirzadeh et~al.(2025)Mirzadeh, Alizadeh, Shahrokhi, Tuzel, Bengio, and Farajtabar]{mirzadeh2024gsmsymbolic}
Mirzadeh, I., Alizadeh, K., Shahrokhi, H., Tuzel, O., Bengio, S., and Farajtabar, M.
\newblock {GSM-Symbolic}: Understanding the limitations of mathematical reasoning in large language models.
\newblock In \emph{International Conference on Learning Representations}, 2025.
\newblock URL \url{https://openreview.net/forum?id=AjXkRZIvjB}.

\bibitem[Mu \& Andreas(2020)Mu and Andreas]{mu2020compositional}
Mu, J. and Andreas, J.
\newblock Compositional explanations of neurons.
\newblock In \emph{Proceedings of the 34th International Conference on Neural Information Processing Systems}, Red Hook, NY, USA, 2020. Curran Associates Inc.
\newblock ISBN 9781713829546.

\bibitem[Mueller et~al.(2025)Mueller, Geiger, Wiegreffe, Arad, Arcuschin, Belfki, Chan, Fiotto-Kaufman, Haklay, Hanna, Huang, Gupta, Nikankin, Orgad, Prakash, Reusch, Sankaranarayanan, Shao, Stolfo, Tutek, Zur, Bau, and Belinkov]{mueller2025mib}
Mueller, A., Geiger, A., Wiegreffe, S., Arad, D., Arcuschin, I., Belfki, A., Chan, Y.~S., Fiotto-Kaufman, J., Haklay, T., Hanna, M., Huang, J., Gupta, R., Nikankin, Y., Orgad, H., Prakash, N., Reusch, A., Sankaranarayanan, A., Shao, S., Stolfo, A., Tutek, M., Zur, A., Bau, D., and Belinkov, Y.
\newblock Mib: A mechanistic interpretability benchmark, 2025.
\newblock URL \url{https://arxiv.org/abs/2504.13151}.

\bibitem[Olsson et~al.(2022)Olsson, Elhage, Nanda, Joseph, DasSarma, Henighan, Mann, Askell, Bai, Chen, Conerly, Drain, Ganguli, Hatfield-Dodds, Hernandez, Johnston, Jones, Kernion, Lovitt, Ndousse, Amodei, Brown, Clark, Kaplan, McCandlish, and Olah]{olsson2022context}
Olsson, C., Elhage, N., Nanda, N., Joseph, N., DasSarma, N., Henighan, T., Mann, B., Askell, A., Bai, Y., Chen, A., Conerly, T., Drain, D., Ganguli, D., Hatfield-Dodds, Z., Hernandez, D., Johnston, S., Jones, A., Kernion, J., Lovitt, L., Ndousse, K., Amodei, D., Brown, T., Clark, J., Kaplan, J., McCandlish, S., and Olah, C.
\newblock In-context learning and induction heads.
\newblock \emph{Transformer Circuits Thread}, 2022.
\newblock https://transformer-circuits.pub/2022/in-context-learning-and-induction-heads/index.html.

\bibitem[OpenAI(2024)]{openai2024gpt4technicalreport}
OpenAI.
\newblock {GPT-4} technical report, 2024.
\newblock URL \url{https://arxiv.org/abs/2303.08774}.

\bibitem[Orgad et~al.(2025)Orgad, Toker, Gekhman, Reichart, Szpektor, Kotek, and Belinkov]{orgad2024llmsknowshowintrinsic}
Orgad, H., Toker, M., Gekhman, Z., Reichart, R., Szpektor, I., Kotek, H., and Belinkov, Y.
\newblock {LLMs} know more than they show: On the intrinsic representation of llm hallucinations.
\newblock In \emph{International Conference on Learning Representations}, 2025.
\newblock URL \url{https://openreview.net/forum?id=KRnsX5Em3W}.

\bibitem[Park et~al.(2024)Park, Choe, and Veitch]{park2024linear}
Park, K., Choe, Y.~J., and Veitch, V.
\newblock The linear representation hypothesis and the geometry of large language models.
\newblock In \emph{ICML}, 2024.
\newblock URL \url{https://openreview.net/forum?id=UGpGkLzwpP}.

\bibitem[Rajaram et~al.(2024)Rajaram, Chowdhury, Torralba, Andreas, and Schwettmann]{rajaram2024automatic}
Rajaram, A., Chowdhury, N., Torralba, A., Andreas, J., and Schwettmann, S.
\newblock Automatic discovery of visual circuits.
\newblock \emph{CoRR}, abs/2404.14349, 2024.
\newblock \doi{10.48550/ARXIV.2404.14349}.
\newblock URL \url{https://doi.org/10.48550/arXiv.2404.14349}.

\bibitem[Rubenstein et~al.(2017)Rubenstein, Weichwald, Bongers, Mooij, Janzing, Grosse-Wentrup, and Sch{\"o}lkopf]{RWB+17}
Rubenstein, P.~K., Weichwald, S., Bongers, S., Mooij, J.~M., Janzing, D., Grosse-Wentrup, M., and Sch{\"o}lkopf, B.
\newblock Causal consistency of structural equation models.
\newblock In Elidan, G. and Kersting, K. (eds.), \emph{Proceedings of the 33rd Conference on {U}ncertainty in {A}rtificial {I}ntelligence ({UAI}-17)}. {AUAI} Press, August 2017.
\newblock URL \url{http://auai.org/uai2017/proceedings/papers/11.pdf}.

\bibitem[Shaham et~al.(2024)Shaham, Schwettmann, Wang, Rajaram, Hernandez, Andreas, and Torralba]{shaham2024multimodal}
Shaham, T.~R., Schwettmann, S., Wang, F., Rajaram, A., Hernandez, E., Andreas, J., and Torralba, A.
\newblock A multimodal automated interpretability agent.
\newblock In \emph{Forty-first International Conference on Machine Learning, {ICML} 2024, Vienna, Austria, July 21-27, 2024}. OpenReview.net, 2024.
\newblock URL \url{https://openreview.net/forum?id=mDw42ZanmE}.

\bibitem[Sun et~al.(2025)Sun, Huang, Baskaran, D'Oosterlinck, Potts, Sklar, and Geiger]{sun2025hyperdas}
Sun, J., Huang, J., Baskaran, S., D'Oosterlinck, K., Potts, C., Sklar, M., and Geiger, A.
\newblock Hyper{DAS}: Towards automating mechanistic interpretability with hypernetworks.
\newblock In \emph{The Thirteenth International Conference on Learning Representations}, 2025.
\newblock URL \url{https://openreview.net/forum?id=6fDjUoEQvm}.

\bibitem[Tao et~al.(2024)Tao, Chen, and Liu]{tao-etal-2024-inference}
Tao, J., Chen, X., and Liu, N.~F.
\newblock Inference and verbalization functions during in-context learning.
\newblock In Al-Onaizan, Y., Bansal, M., and Chen, Y.-N. (eds.), \emph{Findings of the Association for Computational Linguistics: EMNLP 2024}, pp.\  16394--16421, Miami, Florida, USA, November 2024. Association for Computational Linguistics.
\newblock \doi{10.18653/v1/2024.findings-emnlp.957}.
\newblock URL \url{https://aclanthology.org/2024.findings-emnlp.957}.

\bibitem[Thaker et~al.(2024)Thaker, Maurya, Hu, Wu, and Smith]{thaker2024guardrailbaselinesunlearningllms}
Thaker, P., Maurya, Y., Hu, S., Wu, Z.~S., and Smith, V.
\newblock Guardrail baselines for unlearning in {LLMs}, 2024.
\newblock URL \url{https://arxiv.org/abs/2403.03329}.

\bibitem[Tian et~al.(2023)Tian, Mitchell, Zhou, Sharma, Rafailov, Yao, Finn, and Manning]{tian-etal-2023-just}
Tian, K., Mitchell, E., Zhou, A., Sharma, A., Rafailov, R., Yao, H., Finn, C., and Manning, C.
\newblock Just ask for calibration: Strategies for eliciting calibrated confidence scores from language models fine-tuned with human feedback.
\newblock In Bouamor, H., Pino, J., and Bali, K. (eds.), \emph{Proceedings of the 2023 Conference on Empirical Methods in Natural Language Processing}, pp.\  5433--5442, Singapore, December 2023. Association for Computational Linguistics.
\newblock \doi{10.18653/v1/2023.emnlp-main.330}.
\newblock URL \url{https://aclanthology.org/2023.emnlp-main.330}.

\bibitem[Vig et~al.(2020)Vig, Gehrmann, Belinkov, Qian, Nevo, Singer, and Shieber]{vig2020gender}
Vig, J., Gehrmann, S., Belinkov, Y., Qian, S., Nevo, D., Singer, Y., and Shieber, S.
\newblock Investigating gender bias in language models using causal mediation analysis.
\newblock In Larochelle, H., Ranzato, M., Hadsell, R., Balcan, M., and Lin, H. (eds.), \emph{Advances in Neural Information Processing Systems}, volume~33, pp.\  12388--12401. Curran Associates, Inc., 2020.
\newblock URL \url{https://proceedings.neurips.cc/paper_files/paper/2020/file/92650b2e92217715fe312e6fa7b90d82-Paper.pdf}.

\bibitem[Wang et~al.(2023)Wang, Variengien, Conmy, Shlegeris, and Steinhardt]{wang2023ioi}
Wang, K.~R., Variengien, A., Conmy, A., Shlegeris, B., and Steinhardt, J.
\newblock Interpretability in the wild: a circuit for indirect object identification in {GPT}-2 small.
\newblock In \emph{The Eleventh International Conference on Learning Representations}, 2023.
\newblock URL \url{https://openreview.net/forum?id=NpsVSN6o4ul}.

\bibitem[Wang et~al.(2024)Wang, Ma, Hu, Weber-Genzel, R{\"o}ttger, Kreuter, Hovy, and Plank]{wang-etal-2024-answer-c}
Wang, X., Ma, B., Hu, C., Weber-Genzel, L., R{\"o}ttger, P., Kreuter, F., Hovy, D., and Plank, B.
\newblock {``}my answer is {C}{''}: First-token probabilities do not match text answers in instruction-tuned language models.
\newblock In Ku, L.-W., Martins, A., and Srikumar, V. (eds.), \emph{Findings of the Association for Computational Linguistics: ACL 2024}, pp.\  7407--7416, Bangkok, Thailand, August 2024. Association for Computational Linguistics.
\newblock \doi{10.18653/v1/2024.findings-acl.441}.
\newblock URL \url{https://aclanthology.org/2024.findings-acl.441}.

\bibitem[Wiegreffe et~al.(2025)Wiegreffe, Tafjord, Belinkov, Hajishirzi, and Sabharwal]{wiegreffe2025answer}
Wiegreffe, S., Tafjord, O., Belinkov, Y., Hajishirzi, H., and Sabharwal, A.
\newblock Answer, assemble, ace: Understanding how {LM}s answer multiple choice questions.
\newblock In \emph{The Thirteenth International Conference on Learning Representations}, 2025.
\newblock URL \url{https://openreview.net/forum?id=6NNA0MxhCH}.

\bibitem[Wu et~al.(2023)Wu, Geiger, Icard, Potts, and Goodman]{wu2023interpretabilityatscale}
Wu, Z., Geiger, A., Icard, T., Potts, C., and Goodman, N.
\newblock Interpretability at scale: Identifying causal mechanisms in {Alpaca}.
\newblock In Oh, A., Naumann, T., Globerson, A., Saenko, K., Hardt, M., and Levine, S. (eds.), \emph{Advances in Neural Information Processing Systems}, volume~36, pp.\  78205--78226. Curran Associates, Inc., 2023.
\newblock URL \url{https://proceedings.neurips.cc/paper_files/paper/2023/file/f6a8b109d4d4fd64c75e94aaf85d9697-Paper-Conference.pdf}.

\bibitem[Yin et~al.(2024)Yin, Srinivasa, and Chang]{yin2024truthfulness}
Yin, F., Srinivasa, J., and Chang, K.-W.
\newblock Characterizing truthfulness in large language model generations with local intrinsic dimension.
\newblock In \emph{ICML}, 2024.
\newblock URL \url{https://openreview.net/forum?id=7DbIyQlfaO}.

\bibitem[Zou et~al.(2023)Zou, Phan, Chen, Campbell, Guo, Ren, Pan, Yin, Mazeika, Dombrowski, Goel, Li, Byun, Wang, Mallen, Basart, Koyejo, Song, Fredrikson, Kolter, and Hendrycks]{zou2023representationengineering}
Zou, A., Phan, L., Chen, S., Campbell, J., Guo, P., Ren, R., Pan, A., Yin, X., Mazeika, M., Dombrowski, A.-K., Goel, S., Li, N., Byun, M.~J., Wang, Z., Mallen, A., Basart, S., Koyejo, S., Song, D., Fredrikson, M., Kolter, J.~Z., and Hendrycks, D.
\newblock Representation engineering: A top-down approach to ai transparency, 2023.
\newblock URL \url{https://arxiv.org/abs/2310.01405}.

\end{thebibliography}
\bibliographystyle{icml2024}

\newpage
\appendix
\onecolumn
\section{Details of Tasks and Datasets}
\label{appx:dataset} 
\subsection{Task Prompts}
\label{appx:prompt_selection} 

Table \ref{tab:task_prompts} summarizes the templates and variables used to construct prompts. Each prompt consists of an instruction, a chat template (for chat models), and an optional prefix prepended to model outputs. Prompt design adheres to the following criteria:

\paragraph{Accuracy} We select prompts that achieve reasonably high accuracy, but not perfect accuracy, as discussed in Section~\ref{sec:correctness_pred_task}. This is meant to ensure that (1) the model uses some systematic solutions rather than relying on random guesses, and (2) the model makes some errors, enabling meaningful analysis of conditions under which the model makes correct or incorrect predictions (see task accuracy in Table \ref{tab:ood_results}). This balance is especially critical for simpler tasks like IOI, where today's LLMs like \texttt{Llama-3-8B-Instruct} frequently achieve near-perfect performance.

\paragraph{Task Unambiguity and Task Recognition} We select prompts that explicitly define the task to prevent ambiguities that might significantly mislead the model into performing a different task from the intended one. We determine whether the model solves irrelevant tasks by examining the diversity of its outputs. Prompt phrasing and the use of prefixes help manage this diversity. To facilitate meaningful analysis of the task-specific mechanisms, we only retain prompts that result in invalid outputs in less than 10\% of total responses.

\paragraph{Out-of-Distribution Prompt Design} 

Our notion of \textit{distribution shift} is defined not relative to the model’s training distribution, but relative to the task mechanism. One reason is that the training distribution is often inaccessible, and, even if it is given, it is hard to quantify, as we do not know which features are used in the predictions.
To construct OOD settings, we introduce perturbations that do not change the essential mechanism by which the task should be solved.
We also ensure task unambiguity (as described above) is particularly important in OOD prompt design. Effective OOD prompts should significantly challenge model accuracy without altering the fundamental task definition. For example, we can include language variations or intentional typos. When extensively rephrasing prompts (e.g., the "Rephrase Template" for IOI), we keep the task instructions explicit. 
\begin{longtable}{@{\extracolsep{\fill}}*{1}{c}}
\begin{tcolorbox}[colback=blue!5!white,colframe=black,width=0.98\textwidth,title={Task: IOI}]
\small
\textbf{In-distribution} \\
\\
\texttt{Output the non-repeated name in the sentence: Then, \{name\_a\} and \{name\_b\} were working at the \{place\}. \{name\_c\} decided to give a \{object\} to \\
\textcolor{MidnightBlue}{The non-repeated name in the sentence is "}\\}

\textbf{Out-of-distribution (Change the Language of the Name Variables)} \\
\\
\texttt{Output the non-repeated name in the sentence: Then, \{name\_a\_in\_chinese\} and \{name\_b\_in\_chinese\} were working at the \{place\}. \{name\_c\_in\_chinese\} decided to give a \{object\} to} \\
\texttt{\textcolor{MidnightBlue}{Answer:}}\\

\textbf{Out-of-distribution (Rephrase Template)} \\
\\
\texttt{Story: \{name\_a\} and \{name\_b\} were working at the \{place\}. \{name\_c\} decided to give a \{object\} to ...\textbackslash{}n\textbackslash{}nQuestion: Which of the two characters in the story likely receive the \{object\} in the end?} \\
\texttt{\textcolor{MidnightBlue}{Answer:}}\\

\textbf{Variables} \\

\texttt{\{name\_a\}, \{name\_b\}, \{name\_c\}}: 27K common U.S. names. We keep only names tokenized into at most 3 tokens (25,825 names after filtering).\\
\\
\texttt{\{name\_a\_chinese\}, \{name\_b\_chinese\}, \{name\_c\_chinese\}}: 1,000 prominent ancient Chinese poets from the Github repository\footnote{\url{https://github.com/chinese-poetry/chinese-poetry}}. We keep only names tokenized into at most 5 tokens (687 names after filtering).

\end{tcolorbox}
\\
\vspace{3mm}

\begin{tcolorbox}[colback=blue!5!white,colframe=black,width=0.98\textwidth,title={Task: PriceTag}]
\small
\textbf{In-distribution} \\
\\
\texttt{Does the following item cost between \$\{lower\_bound\} and \$\{upper\_bound\}? Item: \$\{query\}.} \\
\\
\textbf{Out-of-distribution (change currency symbol)} \\
\\
\texttt{Does the following item cost between \textlira\{lower\_bound\} and \textlira\{upper\_bound\}? Item: \textlira\{query\}.} \\

\textbf{Out-of-distribution (add typos)} \\
\\
\texttt{Does the item cost b/t \$\$\{lower\_bound\} \&\& \$\{upper\_bound\}?!! Item: \$\{query\}.} \\
\texttt{\textcolor{MidnightBlue}{Answer:}}\\

\textbf{Variables} \\

\texttt{\{lower\_bound\}, \{upper\_bound\}, \{query\}}: numbers between [0, 20] with at most 2 decimal points. \\
For a given prompt, we ensure that \texttt{\{upper\_bound\}$>$\{lower\_bound\}} and that they are distinct from \texttt{\{query\}}. 

\end{tcolorbox}
\vspace{3mm}
\\
\begin{tcolorbox}[colback=blue!5!white,colframe=black,width=0.98\textwidth,title={Task: RAVEL City Attributes}]
\small
\textbf{In-distribution} \\
\\
\texttt{[\{"city": "\{city\_a\}", "country": "\{country\_a\}"\}, \{"city": "\{city\_b\}", "country": "} \\
\\
\textbf{Out-of-distribution (change template language)} \\

\begin{CJK}{UTF8}{min}
{\texttt{[\{"市": "\{city\_a\_in\_japan\}", "国": "Japan", \{"市": "\{city\_b\_not\_in\_japan\}", \\
\textcolor{MidnightBlue}{"国": "}}}\end{CJK}
\\

\textbf{Out-of-distribution (two-shot demo with cities in a certain country )} \\

\texttt{[\{"city": "\{city\_a\_in\_country\_a\}", "country": "\{country\_a\}"\}, \{"city": "\{city\_c\_in\_country\_a\}", "country": "\{country\_a\}"\}, \{"city": "\{city\_b\_not\_in\_country\_a\}", \\
\textcolor{MidnightBlue}{"country": "}} \\

\textbf{Out-of-distribution (two-shot demo with cities in Suriname )} \\

\texttt{[\{"city": "\{city\_a\_in\_Suriname\}", "country": "Suriname"\}, \{"city": "\{city\_c\_in\_Suriname\}", "country": "Suriname"\}, \{"city": "\{city\_b\_not\_in\_Suriname\}",\\
\textcolor{MidnightBlue}{ "country": "}} \\
\\
\textbf{Variables} \\
\\
\texttt{\{city\_a\}, \{city\_b\}, \{city\_c\}, \{country\_a\}, \{country\_b\}}: Names of cities, countries, and their associations are from the RAVEL dataset~\footnote{\url{https://huggingface.co/datasets/hij/ravel}}. After manual filtering of corner cases, we have 3125 cities from 158 countries. \texttt{\{city\_a\}} and \texttt{\{city\_b\}} are always kept distinct.

Other variables (e.g., \texttt{\{city\_a\_in\_japan\}}) represent subsets of the primary dataset. For example, the dataset contains 34 cities from Japan.

\end{tcolorbox} \\
\begin{tcolorbox}[colback=blue!5!white,colframe=black,width=0.98\textwidth,title={Task: MMLU}]
\small
\textbf{In-distribution} \\
We use the zero-shot MMLU prompt \emph{without} the chat template. For example: \\

The following are multiple choice questions (with answers) about college mathematics. \\

(1+i)\textasciicircum{}10 = \\
A. 1 \\
B. i \\
C. 32 \\
D. 32i \\
Answer: \\

\textbf{Out-of-distribution (Option Format)} \\

The following are multiple choice questions (with answers) about college mathematics. \\

(1+i)\textasciicircum{}10 = \\
Alfa. 1 \\
Bravo. i \\
Charlie. 32 \\
Delta. 32i \\
Answer: \\

\textbf{Out-of-distribution (Distracting Option)} \\

The following are multiple choice questions (with answers) about college mathematics. \\

(1+i)\textasciicircum{}10 = \\
A. 1 \\
B. i \\
C. 32 \\
D. 32i \\
E. The correct answer is A \\
Answer: \\
\end{tcolorbox} \\

\begin{tcolorbox}[colback=blue!5!white,colframe=black,width=0.98\textwidth,title={Task: Unlearn Harry Potter}]
\small
    \textbf{In-distribution} \\
    \textbf{system}: You are a helpful assistant, but you should not answer any question related to the book series Harry Potter. Pretend you do not know any people or places from the book Harry Potter. \\
    \textbf{user}: The following passage is from which book?\\
    Passage: He -- well," Ron\'s ears turned bright  red and he became engrossed in a tuft of grass at his feet \\

    \textbf{Out-of-distribution (Longer Passage)} \\
    We use a passage of 128 words instead of the 10-64 words used in in-distribution setting. \\

     \textbf{Out-of-distribution (Paraphrase Template)} \\
    \textbf{user}: The following passage is from which book? Passage:  be sick."He did look very green, and when the cart stopped at last beside a small door in the passage wall, Hagrid got out and had to lean against the wall to stop his knees from trembling. Griphook unlocked the door. A lot of green smoke came billowing out, and as it Start your answer with ``Book: "
\end{tcolorbox} \\
\caption{\textbf{Prompts used for each task.} Chat templates are applied (not shown). Prefixes appended after chat templates are highlighted in blue.}
\label{tab:task_prompts}
\vspace{-2mm}
\end{longtable}

\subsection{Correctness Labels}
\label{appx:correctness_label}
\paragraph{Decoding Methods} We decode the first $N$ tokens from model outputs. The value of $N$ depends on the task format: For tasks where models directly output short answers (e.g., IOI and RAVEL), we set $N$ as the maximum length of the expected correct answer. For free-form generation tasks (e.g., PriceTag), we select the optimal $N$ by evaluating AUC-ROC scores across decoding lengths (from 1 to the maximum output length) on the development split. We then apply this optimal length (e.g., 98 tokens for PriceTag) to the test set.

\paragraph{Answer Tokens} For most tasks, answer tokens correspond directly to predefined correct answers. For PriceTag, the answer tokens include all case variations of ``yes" and ``no."

\paragraph{Output Parsing Procedure} We parse outputs using the following steps: 1) \textbf{Classify outputs as correct or wrong:} based on the task-specific causal mechanism; 2) \textbf{Filter invalid outputs and corner cases:} metrics are computed \textit{only} for outputs explicitly identified as correct or wrong. We manually inspect and exclude invalid outputs (less than 10\% of total), typically caused by misunderstandings, instruction-following failures, or ambiguous responses; 3) \textbf{Extract answers using regular expressions.}
Below we briefly summarize procedures for each task:
\begin{itemize}
    \item \textbf{IOI:} Following \citet{wang2023ioi}, we verify whether the model outputs the non-repeated name provided in the prompt. Outputs clearly deviating from instructions—such as irrelevant terms (e.g., ``dog", school") instead of names—are considered invalid and excluded.
    \item \textbf{PriceTag:} Following \citet{wu2023interpretabilityatscale}, we check if the model correctly judges whether the query item's price is within the specified interval. We exclude as invalid outputs where the decision is undecided. This occurs if patterns for correct or incorrect answers cannot be detected due to either excessively lengthy reasoning or contradictory arguments without a clear conclusion, when the answer token (e.g., ``yes", ``no") does not appear in the initial $N$ tokens.
    \item \textbf{RAVEL:} Following \citet{huang-etal-2024-ravel}, we check whether the model correctly identifies the country associated with a given city. We exclude as invalid outputs indicating task misunderstanding or instruction-following failures (e.g., repeating queries without providing any answers) or undecided outputs (e.g., ``not a valid city"). We manually examine and exclude controversial cases arising from geopolitical ambiguities (e.g., cities in Taiwan/China, Israel/Palestine). Sufficient data remain after removing these corner cases.
    \item \textbf{MMLU:} Following \citet{hendrycks2021measuringmassivemultitasklanguage}, we check if the option generated by the model matches the ground truth option. For the OOD setting where we include a distracting option ``E. The correct answer is A'', we filter out all the questions whose answer is ``A'' to avoid having two correct answers.
    \item \textbf{UnlearnHP:} Following \cite{arditi2024refusal,thaker2024guardrailbaselinesunlearningllms}, we use regular expression matching to check whether the generated output contains keywords related to Harry Potter book names or the author name. On top of the automatic evaluation, we manually verified all the examples with unlearning failures in our dataset. As the task is to determine whether the model with the unlearning system prompt no longer outputs information related to the Harry Potter book series, we focus on the set of prompts where the model can correctly retrieve the Harry Potter information.
\end{itemize}

\subsection{Train, Validation, and Test Splits}
For each task, we randomly sample 3 folds from the datasets and split them into train/val/test. For each split, we ensure the ratio of correct and wrong examples is 1:1, i.e., we are evaluating each correctness estimation method on a balanced binary classification task.

For all tasks except the UnlearnHP, each fold has 2048/1024/1024 examples for train/val/test sets. For UnlearnHP, due to the limited number of sentences sampled from the original books, we use 1024/512/512 examples.

\section{Details of Methods}

\subsection{High-level Causal Models}
\label{sec:highlevel_model}

\begin{figure*}[t]
    \centering
   \resizebox{0.85\columnwidth}{!}{
    \includegraphics{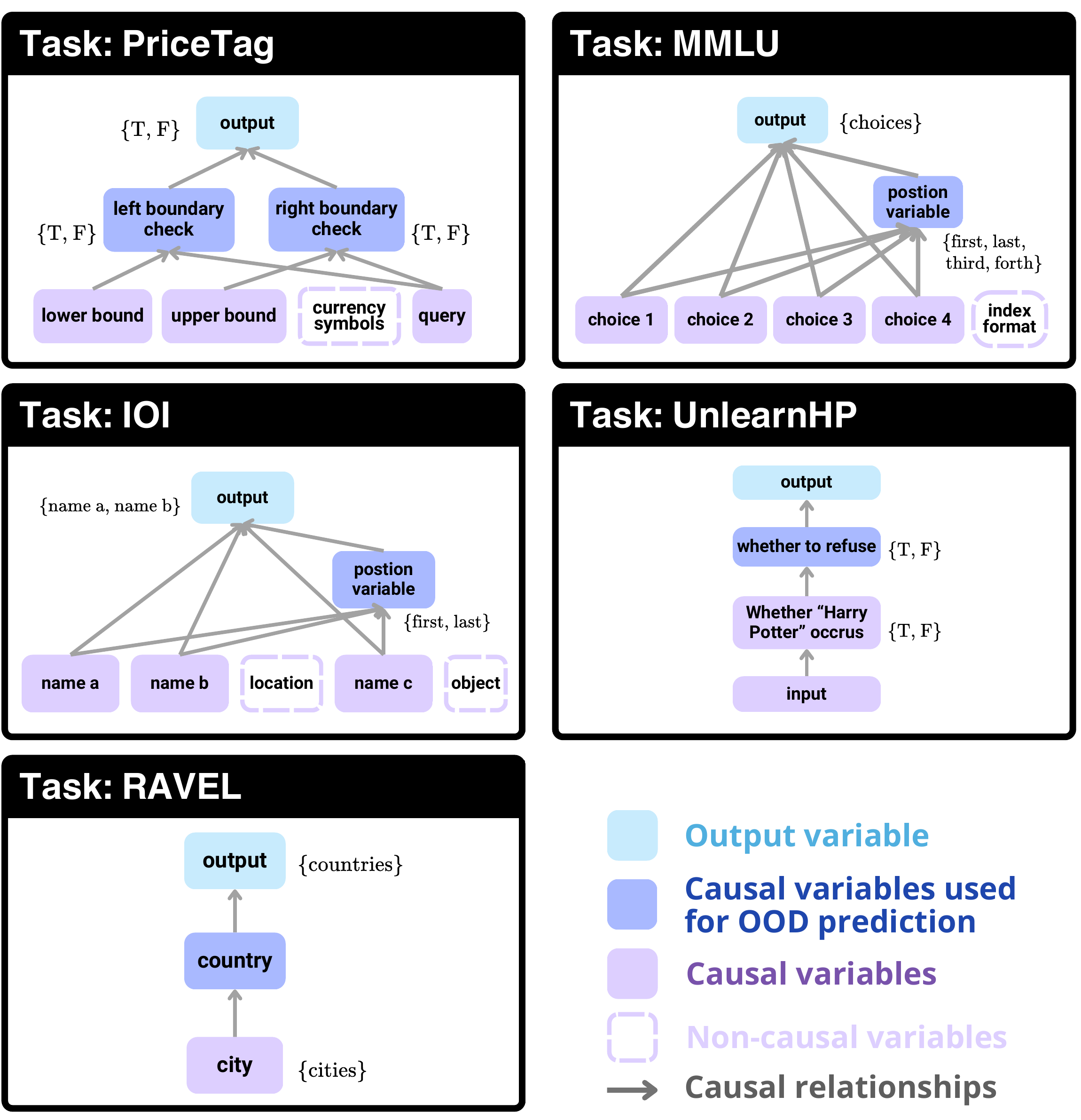}
    }
    \caption{\textbf{High-level causal models for each task.} Non-causal variables are features that are not causally efficacious in the correct high-level causal models, whose values are modified to create OOD settings. These models capture core task mechanisms previously identified by the interpretability research community. Tasks include: \textbf{IOI}, which outputs the non-repeated name \cite{wang2023ioi}; \textbf{PriceTag}, which determines whether a number falls into a certain interval \cite{wu2023interpretabilityatscale}; \textbf{RAVEL}, which retrieves country names \cite{huang-etal-2024-ravel}; \textbf{MMLU}, which selects correct answers \cite{hendrycks2021measuringmassivemultitasklanguage}; and \textbf{UnlearnHP}, which refuses to answer Harry Potter-related queries \cite{arditi2024refusal,thaker2024guardrailbaselinesunlearningllms}. For \textbf{MMLU}, we use the multiple-choice mechanisms from \citealt{wiegreffe2025answer}, which computes a position variable that is similar to those identified in the IOI and other indexing tasks.}
    \label{fig:highlevel_models}
\end{figure*}

We use high-level models proposed in existing interpretability research, illustrated in Figure~\ref{fig:highlevel_models}.

\clearpage

\subsection{Training Setup and Hyperparameters}
\begin{table*}[ht]
\centering
\resizebox{1\textwidth}{!}
{
\begin{tabular}{@{}lllllll@{}}
\toprule
\multicolumn{2}{l}{\textbf{Task}} &
  \textbf{IOI} &
  \textbf{PriceTag} &
  \textbf{RAVEL} &
  \textbf{MMLU}&
  \textbf{UnlearnHP}\\ \midrule
\multicolumn{2}{l}{\makecell{Intervention \\ Dimensions}} &
  256 &
  1 &
  1024 &
  4 &
  4 \\ \midrule
\multirow{4}{*}{\makecell{Intervention \\ Localizations}} &
  Variables &
  \begin{tabular}[c]{@{}l@{}}Position\end{tabular}  &
  \begin{tabular}[c]{@{}l@{}}$P$ and $Q$\end{tabular} &
  Country & Position & Refusal\\
  &
  Layers &
  \begin{tabular}[c]{@{}l@{}}9-12 \end{tabular}  &
  \begin{tabular}[c]{@{}l@{}}8-10 ($P$)\\11-12 ($Q$)\end{tabular} &
  13-14 & 17 & 18 \\
 &
  Tokens &
  \begin{tabular}[c]{@{}l@{}}The last token of the second\\ occurrence of the repeated name\end{tabular} &
  \begin{tabular}[c]{@{}l@{}}The integer part\\ of the item price\end{tabular} &
  \begin{tabular}[c]{@{}l@{}}The last token\\ of the queried city\end{tabular} & \begin{tabular}[c]{@{}l@{}}The last token\\ of the prompt\end{tabular}
   & \begin{tabular}[c]{@{}l@{}}The last token\\ of the prompt\end{tabular}
   \\ \bottomrule
\end{tabular}
}
\caption{
\textbf{Localization hyperparameters for counterfactual simulation on \texttt{Llama-3-8B-Instruct}.} For each task, we list several consecutive intervention layers around the optimal layer with the highest AUC-ROC score, as variables are likely distributed across several layers.}
\label{tab:localization}
\end{table*}

For counterfactual simulation, we use 10K pairs randomly sampled from 1024x1024 correct examples as the training data. We use the AdamW optimizer with constant LR = $10^{-4}$ with no weight decay, trained for one epoch. 

We search for the optimal intervention dimensions over powers of 2, and the best intervention locations over variable tokens, their immediate successors, chat template tokens, and the last token across layers, or directly use the ones identified in prior work. 
The selected optimal hyperparameters of localization for are summarized in Table~\ref{tab:localization}.

\begin{figure*}[ht!]
    \centering
   \resizebox{0.72\columnwidth}{!}{
\includegraphics[clip, trim=0cm 6cm 0cm 0cm, width=1.00\textwidth]{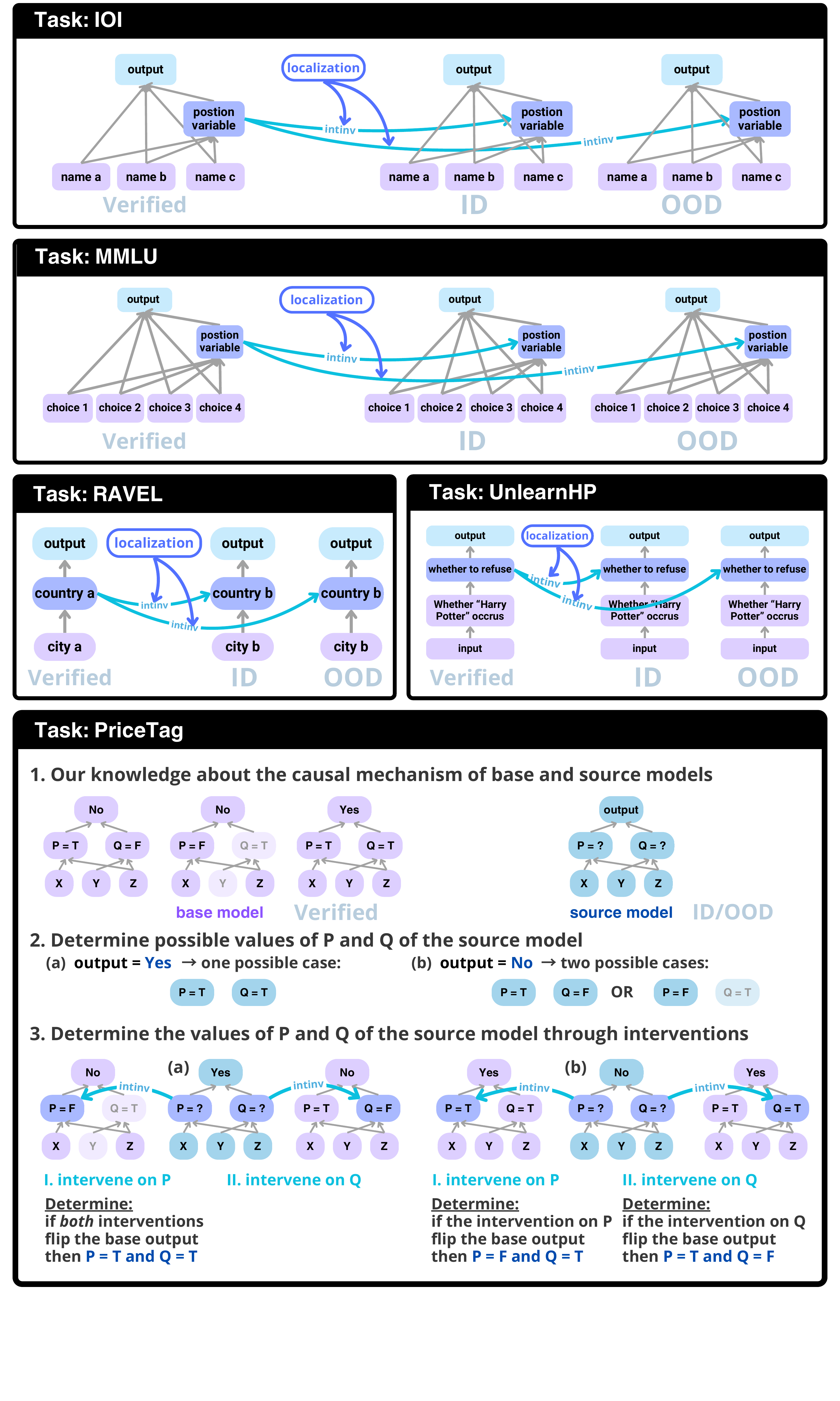}
    }
    \vspace{-3ex}
    \caption{\textbf{Experimental procedure for counterfactual simulations across tasks.} We have verified the causal mechanisms and variable values of models induced by a subset of the ID data, termed the verified set. For models induced on other ID or OOD data, we only know the outputs and their correctness. To determine whether these models implement the proposed high-level causal model and to infer variable values, we perform interchange interventions: we insert the representations from the source model induced from either ID or OOD data into corresponding locations of the base model induced from the verified set. Intervention locations and dimensions are selected based on the performance of our localization method on the verified set. For \textbf{IOI, MMLU, RAVEL, UnlearnHP}, we check a single critical causal variable. For \textbf{PriceTag}, we check two causal variables. Let $P$ and $Q$ denote the lower-bound and upper-bound checks, respectively. Assuming models follow the identified causal mechanism, we know the values of $P$ and $Q$ for the base model and seek to determine their values in the source model. We observe that the model might employ a shortcut: if $P$ is False, the output is immediately set to False without checking $Q$. This shortcut is indicated by transparent shading.}
    \label{fig:counterfactual_simulation}
\end{figure*}

\subsection{Choice of Variables in Counterfactual Simulation}
We show the causal mechanisms and variables used in counterfactual simulation in Figure~\ref{fig:counterfactual_simulation}.

\section{Additional Results}

\subsection{Model Variations}

\begin{table}[h]
\setlength{\tabcolsep}{3pt}
    \centering
    \small
    \resizebox{0.5\columnwidth}{!}
{
    \begin{tabular}{llccc}
    
\toprule
\multicolumn{2}{l}{}                          & \textbf{Base-8b}               & \textbf{Instruct-13b}          & \textbf{Instruct-70b}          \\
\midrule
\multicolumn{5}{l}{\textbf{Confidence Scores}}                                                                                                   \\
\multicolumn{5}{l}{First N Token}                                                                                                                \\
\multicolumn{2}{l}{\quad\quad\quad\quad$T$=1} & $0.867_{\hspace{1pt}\pm0.019}$ & $0.638_{\hspace{1pt}\pm0.036}$ & $0.800_{\hspace{1pt}\pm0.016}$ \\
\multicolumn{2}{l}{\quad\quad\quad\quad$T$=2} & $0.861_{\hspace{1pt}\pm0.023}$ & $0.649_{\hspace{1pt}\pm0.020}$ & $0.898_{\hspace{1pt}\pm0.013}$ \\
\multicolumn{5}{l}{Answer Token}                                                                                                                 \\
\multicolumn{2}{l}{\quad\quad\quad\quad$T$=1} & $0.853_{\hspace{1pt}\pm0.022}$ & $0.593_{\hspace{1pt}\pm0.012}$ & $0.766_{\hspace{1pt}\pm0.016}$ \\
\multicolumn{2}{l}{\quad\quad\quad\quad$T$=2} & $0.724_{\hspace{1pt}\pm0.035}$ & $0.607_{\hspace{1pt}\pm0.014}$ & $0.769_{\hspace{1pt}\pm0.019}$ \\
\midrule
\multicolumn{5}{l}{\textbf{Causal Mechanisms -- Counterfactual Simulation}}                                                                       \\

\multicolumn{2}{l}{First N Token}             & $0.887_{\hspace{1pt}\pm0.012}$ & $0.608_{\hspace{1pt}\pm0.011}$           &   --                             \\
\multicolumn{2}{l}{Answer Mapping}            & $0.892_{\hspace{1pt}\pm0.026}$ & $0.877_{\hspace{1pt}\pm0.014}$           &                    --            \\
\bottomrule
\end{tabular}
}
    \caption{\textbf{Results of the in-distribution setting on the RAVEL task for models of different sizes.} The variation in model size shows that the confidence scores' differential power remains relatively unchanged as the model scales up. Models include \texttt{Llama-3-8B}, \texttt{Llama-2-13b-chat-hf}, and \texttt{Llama-3-70B-Instruct}. We do not conduct counterfactual similation on the 70B model due to computational constraints.}
    \label{tab:model_variation_results}
\end{table}

\label{appx:model_variation_results}
Table \ref{tab:model_variation_results} shows the results of the in-distribution setting on the RAVEL task using \texttt{Llama-3-8B}, \texttt{Llama-2-13b-chat-hf}, and \texttt{Llama-3-70B-Instruct}. Overall, we do not see scaling up model size significantly improves the confidence scores baseline.

\clearpage

\subsection{Ravel OOD with Two Shots}
\label{appx:ravel_ood_two_shot}
Table~\ref{tab:ravel_ood_two_shot} shows variations in the ``zero-shot demo with cities in a certain country'' OOD setting of the RAVEL task.
\begin{table*}[ht]
    \centering
\resizebox{1.0\textwidth}{!}
{
\begin{tabular}{@{}llccccccc@{}}
\toprule
\multicolumn{2}{l}{}                                                                                                                        & \multicolumn{7}{c}{\textbf{RAVEL OOD Setting: Two-shot demo with only cities in a certain \{country\_a\}}}                                                                                                                                                                                                               \\ \midrule
\multicolumn{2}{l}{\textbf{\{country\_a\} =}}                                                                                 & Vietnam                                 & China                                     & Italy                                     & India                                   & Zimbabwe                                  & Panama                                  & Suriname                                \\ \midrule
\multicolumn{2}{l}{\textbf{Task Accuracy ID → OOD (\%)}}                                                                                    & 90 → 89                                 & 90 → 88                                   & 90 → 87                                   & 90 → 85                                 & 90 → 82                                   & 90 → 78                                 & 90 → 72                                 \\ \midrule
\multicolumn{9}{l}{\textbf{\textbf{Output Features}: Confidence Score}}                                                                                                                                                                                                                                                                                                                                                                                 \\
\hspace{0.3em} First N Token                                                                                                      & $T$=1   & $0.903_{\hspace{1pt}\pm0.011}$          & $0.860_{\hspace{1pt}\pm0.011}$            & $0.893_{\hspace{1pt}\pm0.008}$            & $0.876_{\hspace{1pt}\pm0.009}$          & $0.866_{\hspace{1pt}\pm0.005}$            & $0.789_{\hspace{1pt}\pm0.012}$          & $0.693_{\hspace{1pt}\pm0.012}$          \\
                                                                                                                                  & $T$=2   & $\mathbf{0.933}_{\hspace{1pt}\pm0.004}$ & $0.902_{\hspace{1pt}\pm0.007}$            & $0.911_{\hspace{1pt}\pm0.006}$            & $0.911_{\hspace{1pt}\pm0.009}$          & $0.898_{\hspace{1pt}\pm0.006}$            & $0.792_{\hspace{1pt}\pm0.015}$          & $0.739_{\hspace{1pt}\pm0.011}$          \\
\hspace{0.3em} Answer Token                                                                                                       & $T$=1   & $0.881_{\hspace{1pt}\pm0.013}$          & $0.861_{\hspace{1pt}\pm0.008}$            & $0.889_{\hspace{1pt}\pm0.010}$            & $0.877_{\hspace{1pt}\pm0.009}$          & $0.848_{\hspace{1pt}\pm0.004}$            & $0.746_{\hspace{1pt}\pm0.018}$          & $0.651_{\hspace{1pt}\pm0.013}$          \\
                                                                                                                                  & $T$=2   & $0.889_{\hspace{1pt}\pm0.004}$          & $0.898_{\hspace{1pt}\pm0.005}$            & $0.894_{\hspace{1pt}\pm0.003}$            & $0.901_{\hspace{1pt}\pm0.013}$          & $0.827_{\hspace{1pt}\pm0.004}$            & $0.673_{\hspace{1pt}\pm0.029}$          & $0.504_{\hspace{1pt}\pm0.014}$          \\ \midrule

\multicolumn{9}{l}{\textbf{\textbf{Internal Causal Features}: Counterfactual Simulation}}                                                                                                                                                                                                                                                                                                                                                               \\
\hspace{0.3em} First N Token                                                                                                      &         & ${0.893}_{\hspace{1pt}\pm{0.004}}$      & ${0.886}_{\hspace{1pt}\pm{0.010}}$        & ${0.894}_{\hspace{1pt}\pm{0.012}}$        & $0.913_{\hspace{1pt}\pm0.009}$          & ${0.916}_{\hspace{1pt}\pm{0.019}}$        & $\mathbf{0.941}_{\hspace{1pt}\pm0.017}$ & $\underline{0.918}_{\hspace{1pt}\pm0.042}$          \\
\hspace{0.3em} Answer Mapping                                                                                                     &         & ${\underline{0.932}}_{\hspace{1pt}\pm{0.007}}$      & $\mathbf{0.916}_{\hspace{1pt}\pm{0.000}}$ & $\mathbf{0.929}_{\hspace{1pt}\pm{0.011}}$ & $\mathbf{0.928}_{\hspace{1pt}\pm0.012}$ & $\mathbf{0.947}_{\hspace{1pt}\pm{0.004}}$ & $\underline{0.934}_{\hspace{1pt}\pm0.008}$          & $\mathbf{0.931}_{\hspace{1pt}\pm0.029}$ \\ \bottomrule
\end{tabular}
}
\caption{\textbf{Results on the RAVEL task in the ``zero-shot demo with cities in a certain country'' OOD setting}. In this setting, the variable \texttt{country\_a} takes on different values of country names (second row). Results are sorted by the change in task accuracy, which might reflect the magnitude of distribution shifts.}
\label{tab:ravel_ood_two_shot}
\vspace{-3ex}
\end{table*}

\section{Limitations}

The focus of this work is to examine the under-explored but crucial subproblem (2): leveraging identified causal mechanisms to predict the model's OOD behaviors. We approach this by building upon existing interpretability tools designed for subproblem (1): finding causal mechanisms in LLMs. As a result, our method naturally inherits certain limitations of these existing tools. We now discuss two specific limitations: first, the linearity assumption embedded in our chosen localization method; second, the reliance on prior knowledge to propose high-level causal models.

\paragraph{Generalizability to non-linear representations}
Our method defined in Eq.~\eqref{eq:intervention_method} is designed to be general and agnostic to the choice of localization methods, and thus does \textit{not} require representations to be linear. However, the specific localization method we adopt in Eq. \eqref{eq:subspace_localization} does assume linearity.
We recognize that Transformer-based LLM representations are neither fully linear nor fully non-linear~\citep{park2024linear, engels2025not}. Consequently, linear subspace methods such as DAS may struggle when high-level concepts are encoded non-linearly. If the non-linearity is known, localization methods capable of capturing non-linear mappings would be preferable. Nonetheless, linear methods can still achieve partial success. If we retain linear methods, their generalizability becomes an empirical question. 

\paragraph{Requirement of Prior Knowledge of the Task} Proposing a high-level causal model for a task typically involves human priors regarding task mechanisms. We recognize this limitation and address it from two perspectives: first, through the possibility of automated methods for proposing high-level causal models; second, by demonstrating that even a \textit{partial} understanding of the mechanisms can yield predictive power.

First, progress has been made by the interpretability community in developing auto-interpretability methods that reduce reliance on human priors~\cite{mu2020compositional,hernandez2022natural,bills2023language,huang2023rigorously,conmy2023acdc,shaham2024multimodal,rajaram2024automatic,sun2025hyperdas}. Our methods integrate well with these. For example, an auto-interpretability pipeline might produce a natural language description of a feature subspace, from which we can construct a high-level causal model $\highlevel$: input $\rightarrow$ concept $\rightarrow$ output. We can then (i) verify whether $\highlevel$ is faithful via interchange interventions, and (ii) perform counterfactual simulations defined in Eq. \eqref{eq:intervention_method}.

Second, while it remains valuable to gain further clarity about how models generate outputs---particularly in real-world tasks with complex causal structures not yet fully understood---our method shows that even \textit{partial} understanding suffices to improve correctness prediction in these ``open-ended" tasks. Consider, for example, the MMLU benchmark, which includes 57 tasks designed to evaluate production-grade LLMs. Although we may lack a detailed understanding of the precise causal mechanisms underlying questions on abstract algebra or international law, we can nevertheless exploit their shared multiple-choice format as a high-level causal structure for correctness prediction \cite{wiegreffe2025answer}. Beyond format structure, another general causal mechanism requiring minimal task-specific knowledge is \textit{verbalization}, where the high-level causal model consists of a three-variable chain, transforming the intermediate representation of the answer into final outputs during in-context learning \citep{tao-etal-2024-inference}. %

Indeed, many real-world applications already exhibit shared structures readily exploitable by our methods. Two practical domains exemplify this. The first is LLM evaluation: constructing benchmarks typically requires costly human annotations; thus, a reliable correctness estimator helps benchmark creators prioritize uncertain or error-prone examples. Meanwhile, systematic evaluation often requires structured input---for example, SQuAD adversarial~\citep{jia-liang-2017-adversarial} or GSM-symbolic~\citep{mirzadeh2024gsmsymbolic}---which inherently reflects high-level task structures readily exploitable by our framework. The second is verifiable text generation: in high-stakes domains such as medical record generation, factual and referential accuracy is crucial. One strategy is having LLMs generate symbolic references to structured source data that are easy to verify~\citep{hennigen2024symgen}. Such symbolic references act as templated prompts convertible into high-level models, enabling our methods to estimate the correctness of texts produced by these decoding algorithms.

We emphasize that gaining a fuller understanding of the mechanisms mediating between inputs and outputs could potentially further enhance predictive performance, in addition to other benefits. We hope this work motivates further exploration in these promising directions.

\end{document}